\documentclass[11pt,a4paper]{article}
\usepackage{iftex}
\RequireXeTeX
\usepackage{amsmath,amssymb}
\usepackage{fontspec}
\usepackage{unicode-math}
\usepackage[english]{babel}
\usepackage[margin=2.5cm]{geometry}
\usepackage{graphicx,adjustbox}
\usepackage{array,booktabs,longtable}
\usepackage{enumitem}
\usepackage{fvextra}
\fvset{fontsize=\footnotesize,breaklines=true,breakanywhere=true,breakautoindent=true}
\usepackage{xurl}
\usepackage{tikz}
\usetikzlibrary{arrows.meta,positioning}
\usepackage{titlesec}
\titleformat{\section}{\normalfont\Large\raggedright}{}{0pt}{}
\titleformat{\subsection}{\normalfont\large\raggedright}{}{0pt}{}
\titleformat{\subsubsection}{\normalfont\normalsize\raggedright}{}{0pt}{}
\titleformat{\paragraph}{\normalfont\normalsize\itshape}{}{0pt}{}
\titleformat{\subparagraph}{\normalfont\normalsize\itshape}{}{0pt}{}
\usepackage{tocloft}

\AtBeginDocument{\raggedright}
\AddToHook{cmd/tableofcontents/before}{\clearpage}
\AddToHook{env/abstract/after}{\par\smallskip\noindent\textbf{Keywords:} Digital Identity Optimization (DIO); Ontology of Digital Identity (ODI); Agentic Entity Resolution; commit-boundary authorization; identity-preserving constraints; claim-specific evidence promotion.\par}
\usepackage[hidelinks,unicode]{hyperref}
\hypersetup{pdftitle={Constraining the Commit Boundary: A DIO/ODI-Informed Authorization Guardrail for Agentic Entity Resolution},pdfauthor={Daniel Beránek},pdfsubject={A conceptual authorization architecture for persistent identity-changing mutations in Agentic Entity Resolution},pdfkeywords={Digital Identity Optimization; DIO; Ontology of Digital Identity; ODI; digital identity; identity reconstruction; artificial intelligence; search engines; knowledge graphs; human observers; Agentic Entity Resolution; commit-boundary authorization; identity-changing mutations; identity-preserving constraints; referential continuity; semantic coherence; entity resolution; Reconstruction Failure Modes; Referential Reconstruction Failures; Entity Substitution; Merge Failure; Split Failure; Entity Omission; Semantic Coherence Failures; Semantic Fragmentation; Temporal Reconstruction Failures; Temporal Discontinuity; Temporal Misalignment; Contextual Adequacy Failures; Contextual Displacement}}
\title{Constraining the Commit Boundary: A DIO/ODI-Informed Authorization Guardrail for Agentic Entity Resolution}
\author{Daniel Beránek\\
Independent Researcher, Brno, Czech Republic\\
dan@danielberanek.cz}
\date{November 4, 2026}
\begin{document}
\maketitle
\begin{abstract}
Agentic Entity Resolution (ER) systems iteratively plan actions, retrieve evidence, revise hypotheses, and propose graph mutations. An erroneous identity hypothesis may condition later retrieval and produce apparent confirmation. This paper proposes a Digital Identity Optimization (DIO) and Ontology of Digital Identity (ODI)-informed authorization boundary between evidence available for reasoning and evidence eligible to authorize persistent identity-changing mutations. A domain-governed constraint-binding function maps selected reconstruction-failure distinctions to identity-preserving constraints. An abstract two-phase supervisor permits commitment only when claim-specific authorization-eligible evidence supports every required proposition, relevant conflicts are resolved, scope is valid, and bound state and policy versions remain current. Admission as \texttt{UNVERIFIED} does not confer authorization eligibility; separate claim-specific promotion is required, and insufficient support or unresolved material conflict defers the mutation. The guarantee is conditional on adequate domain policy, sound claim-specific evidence promotion, and faithful enforcement; the specification does not independently establish promotion correctness or semantic truth. The proposed contribution is a DIO/ODI-informed diagnostic-to-constraint organization integrated with an explicit authorization contract for claim-specific evidence promotion and mutation-specific proposition requirements. Scope binding, conflict handling, and commit-time version checks are treated as necessary contract elements, not as novel mechanisms in isolation. This is a conceptual architecture without implementation or empirical validation. It does not prevent erroneous hypotheses or hypothesis-conditioned retrieval; dependency-aware evidence flow remains future research.

\end{abstract}
\tableofcontents
\newpage

\section{1. Introduction}
Entity Resolution (ER) is the data-management task of determining whether distinct records, mentions, or representations refer to the same real-world referent. It supports master data management, knowledge-graph construction, and cross-silo integration in heterogeneous data environments [12, 19]. Traditional ER systems compare candidate pairs or clusters using rules, learned representations, or combinations of both [10, 12, 16, 25]. Entity linking and explainable entity matching address adjacent problems at mention or pair level [2, 3], while pretrained language models and LLM-assisted methods broaden the range of evidence and comparisons available to ER pipelines [15, 16, 20, 25]. This paper addresses a governance question that becomes especially important when resolution is iterative: which evidence may authorize a persistent identity change, and under what conditions may a proposed graph mutation be committed?

Recent work has framed Agentic ER as a sequential decision-making process in which an autonomous agent plans actions, gathers evidence iteratively, invokes tools, revises its beliefs, and selectively defers cases [17]. This creates opportunities for adaptive evidence acquisition, but it also makes later actions dependent on prior intermediate states. A mistaken identity hypothesis can condition a retrieval query; the resulting material may be relevant to that query without independently establishing the identity claim. If the agent then treats the material as confirmation, an initial error may become harder to dislodge. We use \emph{apparent verification} for this possible mechanism. It is an analytical construct, not a claim that iterative retrieval inevitably produces self-confirmation; retrieval-augmented systems have been studied for related vulnerability patterns [23].

The problem is not that an agent must never reason with incomplete or uncertain evidence. Exploratory reasoning often requires provisional hypotheses. The architectural question is whether a hypothesis or observation that is useful for planning should automatically be eligible to authorize a persistent change to an entity's identity. This paper proposes that these statuses should be separated. The agent may retain a provisional interpretation for reasoning, while a protected mutation boundary requires evidence that has passed a distinct, claim-specific authorization process. This guarantee is conditional on adequate domain policy, sound claim-specific evidence promotion, and faithful enforcement at the protected write boundary. The supervisor applies the stated contract to the evidence and validation records it receives; it does not independently establish the semantic correctness of the promotion decision or the underlying identity evidence.

The paper connects this boundary to the Digital Identity Optimization (DIO) and Ontology of Digital Identity (ODI) Reference Model [5, 8, 9]. DIO/ODI supplies a reconstruction-oriented vocabulary and diagnostic distinctions across referential, semantic, temporal, and contextual failure modes [4, 6, 7]. The contribution here is not to claim that provenance tracking, invariant checks, transaction versioning, abstention, or human review are new in isolation. It is to propose an explicit bridge between DIO/ODI reconstruction diagnostics and mutation-specific authorization constraints, with a clear boundary around the property the architecture claims to provide.

The proposed contribution has four parts:

\begin{enumerate}
\item A lifecycle-to-boundary analysis that maps possible Agentic ER failure pathways to the guardrail’s controls and explicitly stated limitations.
\item A domain-governed constraint-binding function that translates selected DIO/ODI distinctions into inspectable identity-preserving conditions for particular mutation classes.
\item An abstract two-phase Minimum Viable Digital Identity Optimization (MVDIO) Supervisor Layer that separates Planner-visible evidence from claim-specific authorization-eligible evidence and gates persistent graph mutations on semantic evaluation, authorized scope, and valid state/policy versions.
\item A prospective evaluation protocol that separates the effect of DIO/ODI-informed constraints from conventional identifier rules, generic authorization gates, and default-deny/abstention baselines.
\end{enumerate}
The formal invariant summarizes these obligations; it is not presented as an independent theorem of semantic correctness. The contribution is the proposed DIO/ODI-informed organization of domain constraints for mutation-specific authorization, integrated with established requirements for claim-specific evidence promotion, proposition coverage, conflict handling, scope binding, and commit-time version integrity. These contract requirements are not claimed as individually novel.

This is a conceptual architecture and research proposal, not a completed implementation or an empirical claim of effectiveness. In particular, it does \textbf{not} prevent the agent from forming a wrong hypothesis, eliminate hypothesis-conditioned retrieval, or establish that all apparent confirmations are detected. A broader information-flow design that tracks dependencies between hypotheses, actions, observations, and subsequent claims remains a separate research direction.

\begin{figure}[htbp]
\centering
\includegraphics[width=\linewidth]{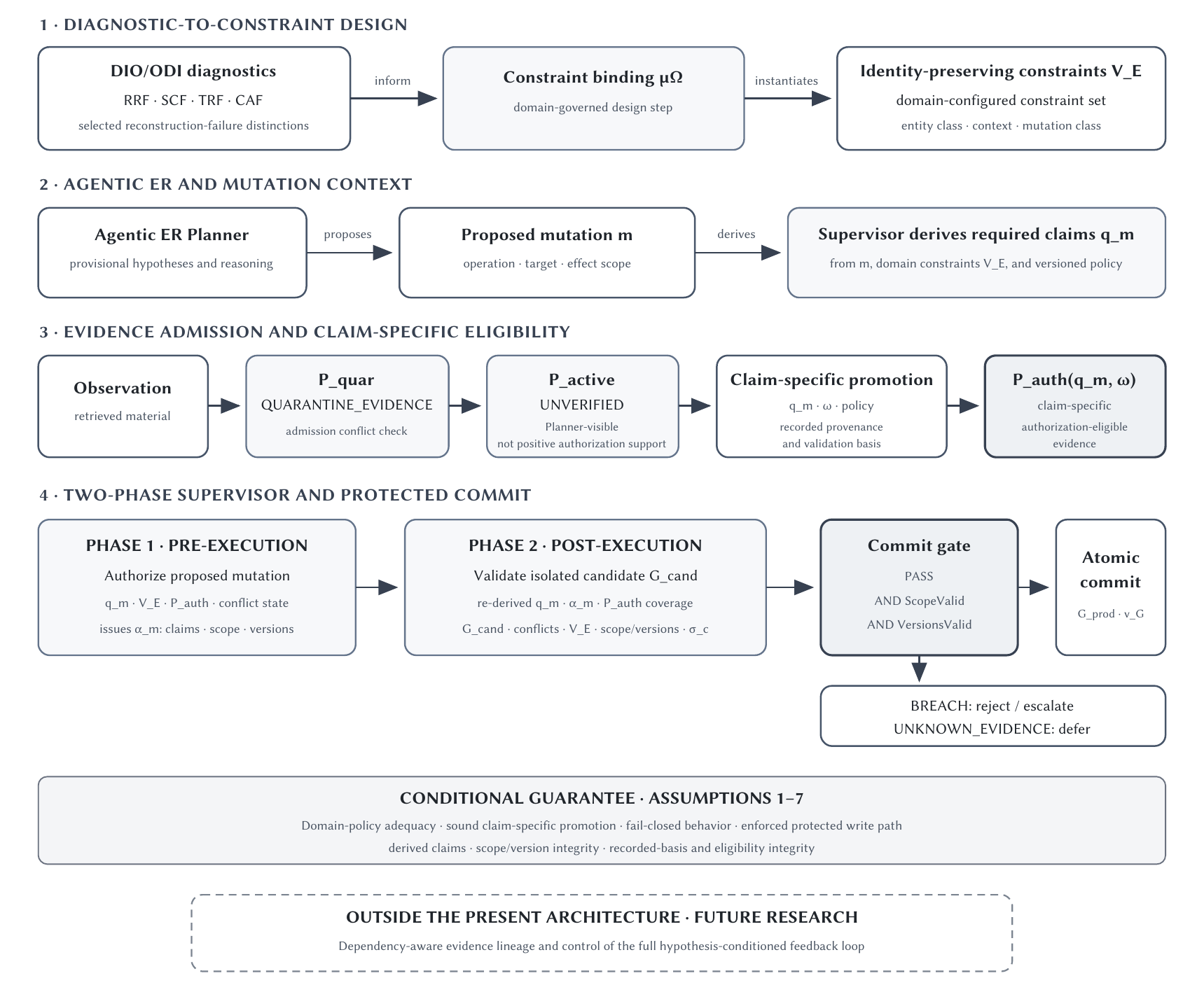}
\caption{Overview of the proposed DIO/ODI-informed commit-boundary architecture. Selected diagnostic distinctions inform domain-configured identity-preserving constraints; the proposed mutation and versioned policy determine required propositions; and observations progress through admission and claim-specific promotion. \texttt{UNVERIFIED} evidence remains available for Planner reasoning but cannot provide positive authorization support, although it may indicate an unresolved conflict. The two-phase supervisor validates the proposed operation and then the realized candidate. A protected commit requires \texttt{PASS}, valid scope, and valid versions at the same commit snapshot. The guarantee is conditional on Assumptions 1--7. Dependency-aware control of the upstream feedback loop is outside the present architecture.}
\label{fig:architecture-overview}
\end{figure}
\section{2. Related Work and Positioning}
The general mechanisms used by the proposed guardrail—authorization gates, provenance records, version checks, graph validation, scope binding, and conservative deferral—are established in adjacent work. The comparison below therefore focuses on each line's relevant contribution and the narrower identity-resolution question addressed here; it does not imply that a mechanism omitted from a row is absent from the cited system.

{\small
\begin{longtable}{@{}>{\raggedright\arraybackslash}p{0.23\dimexpr\linewidth-4\tabcolsep\relax} >{\raggedright\arraybackslash}p{0.34\dimexpr\linewidth-4\tabcolsep\relax} >{\raggedright\arraybackslash}p{0.43\dimexpr\linewidth-4\tabcolsep\relax}@{}}
\caption{Positioning relative to adjacent work.}\\
\toprule
\textbf{Prior work / line} & \textbf{Established contribution relevant here} & \textbf{Scope and distinction of this paper} \\
\midrule
\endfirsthead
\toprule
\textbf{Prior work / line} & \textbf{Established contribution relevant here} & \textbf{Scope and distinction of this paper} \\
\midrule
\endhead
Conventional ER, LLM-assisted ER, and Agentic ER [10, 12, 15–17, 20, 25] & Entity matching, data integration, language-model assistance, and sequential agent workflows. & Does not propose a new matching algorithm or agent workflow. Focuses on authorization conditions for persistent identity-changing mutations. \\
\addlinespace
Constraint-based deduplication and ER [37] & Declarative and domain-specific constraints can restrict collective deduplication; must-link/cannot-link constraints and richer relational constraints are established approaches. & Does not claim to introduce constraint-based ER. Uses selected DIO/ODI reconstruction distinctions to organize mutation-specific propositions and authorization at an agentic persistence boundary. \\
\addlinespace
Ontology-governed enterprise agents [29] & Ontology-grounded entity alignment, graph simulation, isolated staging, provenance-bearing traces, and confidence-gated acceptance with clarification or ontology-gap handling. & Targets heterogeneous ER merges and identity-continuity claims through DIO/ODI-informed, mutation-specific policy rather than a configured enterprise-ontology scenario. \\
\addlinespace
Runtime authorization of agent-generated subgoals [28] & Version-bound authorization witnesses and rechecking protected effects at an atomic commit boundary. & Treats the general runtime authorization boundary as prior work; focuses on the identity or continuity propositions and evidence required for a particular ER mutation. \\
\addlinespace
Commit-time authorization [32] & Freshness, causal precedence, effect binding, and revalidation of authority for durable effects. & Shares the commit-time concern but asks which identity or continuity proposition must be established, not only whether earlier authority remains valid. \\
\addlinespace
Hard-KLAIM and AGATE [33, 34] & Hard-KLAIM: typed effects, operational semantics, proposal–approval–commitment binding, provenance constraints, and mechanized trace-safety proofs. AGATE: an implemented runtime authorization/provenance gate with parameter-bound grants, source observations, effect tracking, and empirical evaluation. & Does not claim comparable formal proof, implementation, or empirical evaluation. Proposes an ER-specific semantic authorization contract. \\
\addlinespace
EBL-Core and CONTINUITY [35, 36] & Candidate-bound execution authorization, evidence and policy binding, current-condition checks, and continuity of authorization context across agent components. & Does not propose another general execution-boundary contract; focuses on identity propositions and domain constraints for an ER mutation. \\
\addlinespace
ARGUS [22] & Influence-provenance tracing, selective admission of context as usable evidence, and evidence-sensitive authorization of consequential actions. & Addresses a different threat model. The present design does not reconstruct comprehensive causal lineage and does not claim to solve the full feedback cascade. \\
\addlinespace
SHACL and PROV-O [30, 31] & Graph-constraint validation and standardized representation of provenance. & Provide relevant concepts or potential implementation substrates, but do not by themselves specify the DIO/ODI diagnostic-to-constraint mapping proposed here. \\
\addlinespace
DIO/ODI reference work [4–9] & Reconstruction-oriented concepts and distinctions covering referential, semantic, temporal, and contextual failure modes. & Supplies the diagnostic vocabulary used in the proposed domain-governed mapping; does not automatically generate the executable constraints or establish their incremental value. \\
\addlinespace
\bottomrule
\end{longtable}
}
Constraint-based ER is also prior art, not an omitted precursor. Must-link/cannot-link constraints and declarative relational constraints, including Dedupalog, already express conditions on deduplication outcomes [37]. At the diagnostic level, the present RRF distinctions overlap with familiar over-merging and under-merging error classes: Merge Failure corresponds to false co-reference or over-merging, while Split Failure corresponds to missed co-reference or under-merging. The DIO/ODI terms retain their broader reconstruction framing; this paper uses them to organize selected domain constraints, not to rename or replace conventional ER error analysis.

The closest authorization-related neighbors establish that version-bound commit checks and evidence-aware execution gates are not, by themselves, novel. Zhu and Wang [28] authorize self-generated subgoals against a principal-approved task contract, while Santos-Grueiro [32] examines whether authority for a durable effect remains fresh, causally prior, and valid at commit time. Zhu et al. [29] additionally cover ontology-grounded entity alignment and scenario-driven graph simulation, with low-confidence cases held from simulation for clarification or ontology-gap resolution. The distinction proposed here is narrower: the required identity or continuity proposition is derived for a particular ER mutation and assessed under domain-configured constraints informed by selected DIO/ODI diagnostics.

Hard-KLAIM [33], AGATE [34], EBL-Core [35], CONTINUITY [36], and ARGUS [22] further establish relevant approaches to typed effects, runtime authorization, provenance-aware execution, current-condition checks, or cross-component authorization continuity. The present manuscript does not claim comparable formal guarantees or an implemented runtime defense. Nor does it claim comprehensive causal-lineage reconstruction; the handling of dependencies that are absent from recorded provenance remains outside the stated guarantee.

SHACL [30] specifies graph-shape validation, and W3C PROV-O [31] provides a vocabulary for representing and exchanging provenance. These can support implementation of parts of an authorization architecture, but neither supplies the policy decision about which identity-changing proposition must be established for a particular mutation. Similarly, the DIO/ODI contribution here is not automatic SKOS entailment or an established advantage over a generic policy built from the same domain facts. It is the proposed diagnostic-to-constraint mapping: selected reconstruction-failure distinctions organize which claim a mutation requires, which relations are insufficient to establish it, and when temporal or contextual conditions call for deferral. Its incremental value remains an open empirical question.

\section{3. Diagnostic Failure-Path Analysis of Agentic ER}
The Agentic ER operational cycle alternates through four stages: \emph{Plan} (select actions from the current state), \emph{Act} (execute a tool or operation), \emph{Observe} (receive evidence or a mutation result), and \emph{Update Beliefs} (revise the internal state) [17]. The following mapping applies the DIO/ODI Reconstruction Failure Mode taxonomy diagnostically to that cycle [7, 8]. It is a proposed analytical application of the taxonomy, not a claim that the taxonomy was part of the original Agentic ER proposal.

The DIO/ODI diagnostic vocabulary covers four dimensions:

\begin{itemize}
\item \textbf{Referential Reconstruction Failures (RRF):} Entity Substitution (ES), Merge Failure (MF), Split Failure (SF), and Entity Omission.
\item \textbf{Semantic Coherence Failures (SCF):} Semantic Fragmentation (SFR).
\item \textbf{Temporal Reconstruction Failures (TRF):} Temporal Discontinuity (TD) and Temporal Misalignment (TM).
\item \textbf{Contextual Adequacy Failures (CAF):} Contextual Displacement (CD).
\end{itemize}
The diagnostic dimensions inform analysis across the Agentic ER lifecycle, but the proposed commit-boundary guardrail operates only at the persistence boundary: it governs mutation authorization and commitment, not all hypothesis formation, query conditioning, or intermediate belief updates.

\subsection{3.1 Referential Reconstruction Failures (RRF)}
Referential failures concern incorrect identity assignment or cardinality violations between an intended referent \(A\) and its reconstructed identity \(\widehat{A}\) [6, 7].

\begin{itemize}
\item \textbf{Merge Failure} (\(\{A,B\}\rightarrow\widehat{X}, A\neq B\)): distinct referents are consolidated into one representation [7, 12].
\item \textbf{Split Failure} (\(A\rightarrow\{\widehat{A}_1,\widehat{A}_2,\ldots\}\)): one referent is represented as disconnected identity clusters because temporal or relational continuity was not reconstructed.
\item \textbf{Entity Substitution} (\(A\rightarrow\widehat{B}, A\neq B\)): a less prominent target is displaced by a structurally or lexically prominent namesake. Entity-overshadowing research provides a relevant example of this ambiguity class [18]; entity-linking work provides adjacent context [2].
\item \textbf{Entity Omission:} the system fails to generate a reconstruction despite sufficient representational grounds within the defined observation domain.
\end{itemize}
In a sequential agent, an early referential failure may influence later query formulation, evidence selection, and mutation proposals. The resulting process might compound the error, correct it, or leave it unchanged. The authorization design proposed here does not determine which trajectory occurs; it governs whether a persistent identity-changing mutation is permitted at the final protected boundary.

\subsection{3.2 Semantic Coherence Failures (SCF)}
\textbf{Semantic Fragmentation} occurs when the system retains referential unity but fails to reconcile incompatible relational claims or representations into a coherent profile [6, 7]. Referential correctness alone does not guarantee that the resulting set of claims is coherent or usable. A mutation that consolidates records must not be treated as resolving contradictions merely because the records have been clustered together.

\subsection{3.3 Temporal Reconstruction Failures (TRF)}
\textbf{Temporal Discontinuity} and \textbf{Temporal Misalignment} arise when representations are not adequately anchored in time [4, 24]. Temporal misalignment assigns an obsolete state to a referent whose identity remains correct; temporal discontinuity treats historical states of a continuous referent as separate entities. This distinction matters for acquisitions, legal renaming, changing officers, product transitions, and other cases where attributes and relations legitimately evolve.

\subsection{3.4 Contextual Adequacy Failures (CAF)}
\textbf{Contextual Displacement} occurs when a reconstruction may be factually valid but does not satisfy the requirements of the current observation domain \(\Omega\) [6, 7]. A matching result adequate for product-catalog deduplication may not be adequate for legal-entity compliance. Domain and purpose therefore matter when translating a diagnostic concept into an authorization rule.

\subsection{3.5 Failure Stage, Control Boundary, and Limitation}
The failure mapping is diagnostic; the following table makes its operational scope explicit.

{\small
\begin{longtable}{@{}>{\raggedright\arraybackslash}p{0.25\dimexpr\linewidth-6\tabcolsep\relax} >{\raggedright\arraybackslash}p{0.25\dimexpr\linewidth-6\tabcolsep\relax} >{\raggedright\arraybackslash}p{0.25\dimexpr\linewidth-6\tabcolsep\relax} >{\raggedright\arraybackslash}p{0.25\dimexpr\linewidth-6\tabcolsep\relax}@{}}
\caption{Failure pathways across the Agentic ER lifecycle and the scope of the proposed commit-boundary control.}\\
\toprule
\textbf{Lifecycle stage} & \textbf{Potential failure pathway} & \textbf{What the commit-boundary guardrail does} & \textbf{What it does not claim to do} \\
\midrule
\endfirsthead
\toprule
\textbf{Lifecycle stage} & \textbf{Potential failure pathway} & \textbf{What the commit-boundary guardrail does} & \textbf{What it does not claim to do} \\
\midrule
\endhead
Plan & A mistaken identity hypothesis conditions an action or query (potential RRF, TRF, or CAF pathway). & Binds any proposed mutation to an explicit authorization record and policy scope. & Does not prevent a wrong hypothesis from shaping the Planner. \\
\addlinespace
Act & A query retrieves material selected or phrased under the current hypothesis (potential CAF pathway). & Checks the declared action scope; retrieval results enter a controlled admission path. & Does not guarantee independent retrieval or infer causal independence. \\
\addlinespace
Observe & Retrieved content appears consistent with the hypothesis or conflicts with other claims (potential SCF/TRF pathway). & Stages observations; records detected material conflicts for deferral/adjudication. & Does not reliably detect all apparent verification or infer causal dependence among sources. \\
\addlinespace
Update Beliefs & An observation affects the Planner's internal state and may reinforce a substitution or stale interpretation (potential RRF/TRF pathway). & Separates Planner eligibility from authorization eligibility. & Does not block all effects of \texttt{UNVERIFIED} material on internal reasoning. \\
\addlinespace
Persistent graph mutation & A hypothesis is converted into a durable merge or identity-changing edge. & Requires a \texttt{PASS} under the declared constraints on claim-specific \(P_{\mathrm{auth}}\), valid scope, and current state/policy versions. & Does not prove that the domain policy or evidence validator is infallible. \\
\addlinespace
\bottomrule
\end{longtable}
}
This table is a diagnostic mapping of possible failure pathways to the proposed control boundary. It does not claim that the lifecycle stages uniquely determine a failure category or that the proposed guardrail detects every listed pathway. The contribution concerns \textbf{authorization containment at the persistence boundary}, conditional on the assumptions specified below, not a full information-flow guarantee.

\section{4. A Stylized Model of Apparent Verification}
An abstract sequential agent may be represented by a state \(s_t\in\mathcal S\), an action \(a_t\in\mathcal A\), the observation \(o_{t+1}\in\mathcal O\) returned after that action, and an update function

\begin{center}
\begin{adjustbox}{max width=\linewidth}
$\displaystyle s_{t+1}=\Phi(s_t,a_t,o_{t+1}).$
\end{adjustbox}
\end{center}
The indexing convention is action-first: executing \(a_t\) from \(s_t\) returns \(o_{t+1}\), which is incorporated into \(s_{t+1}\).

\subsection{4.1 A Possible Hypothesis-Conditioned Feedback Path}
Under hypothesis-conditioned retrieval and insufficiently independent validation, the following sequence is possible:

\begin{enumerate}
\item \textbf{Latent distortion:} the state \(s_t\) contains an erroneous identity hypothesis introduced by a noisy classifier, prior inference, or earlier evidence [7, 17].
\item \textbf{Biased action parameterization:} the next action is selected using the corrupted state, so a query or tool call encodes part of the earlier hypothesis [4, 11].
\item \textbf{Contextual ingestion:} retrieval returns material relevant to the query as formulated, but relevance does not independently establish the target referent's identity [4, 23].
\item \textbf{Apparent convergence:} the apparent discrepancy between an observation and the existing hypothesis may decrease even when the hypothesis is wrong, because the observation process was conditioned on that hypothesis.
\end{enumerate}
A stylized residual is

\begin{center}
\begin{adjustbox}{max width=\linewidth}
$\displaystyle \left\|\operatorname{Project}(o_{t+1})-\operatorname{Expectation}(s_t)\right\|\rightarrow 0.$
\end{adjustbox}
\end{center}
Here, \texttt{Project} denotes an illustrative mapping of an observation into the representation used by the matcher, and \texttt{Expectation} denotes the representation predicted from the current state. The expression presupposes that both outputs are comparable in a common representational space; neither mapping nor the discrepancy measure is specified as a calibrated estimator. The arrow denotes a possible reduction in apparent discrepancy over a feedback sequence, not a convergence result established by this paper. The expression illustrates how a hypothesis-conditioned observation may appear to agree with the current state without independently establishing referential correctness.

\textbf{Boundary of the proposed solution.} The proposed MVDIO Supervisor does not close the Plan–Act–Observe–Update feedback loop. It does not claim that a query is independent of the current hypothesis, that multiple returned sources provide independent corroboration, or that intermediate belief updates are free of contaminated evidence. It addresses the downstream authorization question: whether the evidence and constraints recorded for a proposed persistent mutation are sufficient to permit that mutation. This narrower boundary is deliberate and makes the proposed property inspectable without claiming a solution to the full cascade mechanism.

\section{5. DIO/ODI-Informed Mutation Authorization}
The authorization design uses a reconstruction-oriented semantic model to inform constraints over identity-changing operations. A full agent planner may be modeled as a constrained sequential decision process, such as a Constrained Markov Decision Process [1], but this paper does not specify or claim to solve the Planner's global optimization problem. Its formal focus is the authorization condition at a persistent mutation boundary.

\subsection{5.1 Semantic Target: Reconstruction Admissibility and Observation Context}
Let \(\omega=(O,C,t)\in\Omega\) denote the observation domain relevant to a mutation: an observer or consuming system \(O\), an operational context or purpose \(C\), and the relevant temporal epoch or interval \(t\). Let \(\mathcal R\) denote the space of analytically considered reconstructions. For a referent \(E\), define the ideal admissible reconstruction set as

\begin{center}
\begin{adjustbox}{max width=\linewidth}
$\displaystyle \operatorname{Adm}(E,\omega)\subseteq\mathcal R.$
\end{adjustbox}
\end{center}
Here, \(\operatorname{Adm}(E,\omega)\) denotes the reconstructions that preserve the intended referential assignment and satisfy the applicable semantic and temporal conditions for \(E\) under \(\omega\). It is an ideal semantic reference, not a set that the proposed supervisor can enumerate or verify directly from incomplete evidence.

Let \(\mathcal S\) denote the relevant agent-state space, and let the partial projection from a state to a reconstruction be

\begin{center}
\begin{adjustbox}{max width=\linewidth}
$\displaystyle \Gamma:\mathcal S\rightharpoonup\mathcal R.$
\end{adjustbox}
\end{center}
The partiality matters: some states do not yield a well-defined reconstruction in the observation domain. For states where a reconstruction is defined, the corresponding idealized inadmissibility set is

\begin{center}
\begin{adjustbox}{max width=\linewidth}
$\displaystyle \mathcal S_{\mathrm{ideal\_fail}}(E,\omega):=\{s\in\operatorname{dom}(\Gamma)\mid \Gamma(s)\notin\operatorname{Adm}(E,\omega)\}.$
\end{adjustbox}
\end{center}
This set is an analytical reference for a defined but inadmissible reconstruction; it is not an executable detector or a claim that the supervisor observes the agent's full internal state. The expression does not classify an undefined reconstruction and must not be read as proof that an identity claim is false. Entity Omission—where no reconstruction is produced despite sufficient grounds within the observation domain—and upstream acquisition/coverage failures require separate analysis; the latter should not automatically be labeled reconstruction failures.

The diagnostic distinctions in Section 3 organize different ways in which reconstruction can fail or become unsuitable. Defined-output cases of Entity Substitution, Merge Failure, and Split Failure can be characterized as candidate cases of idealized inadmissibility when assessed for the relevant referent and observation domain. The set does not itself classify the cause of inadmissibility or provide a one-to-one mapping to RRF, SCF, TRF, and CAF. In particular, Semantic Fragmentation is relational across multiple reconstructions, while Temporal Discontinuity may require a sequence of time-indexed states; neither is fully encoded by this single-state set definition. The taxonomy in Section 3 remains a diagnostic organization, not a partition theorem over \(\mathcal S_{\mathrm{ideal\_fail}}\). Contextual Displacement is analytically distinct: a reconstruction may preserve referential identity while remaining inadequate for the current purpose, so contextual conditions must be represented by domain policy rather than inferred from referential inadmissibility alone.

Contextual fitness remains distinct from admissibility. A reconstruction may preserve referential identity and temporal continuity yet still be inadequate for a particular task or consuming system. We do not assign a numeric contextual-adequacy score here, because no calibrated scoring function is specified. Instead, context-sensitive requirements must be expressed as explicit, versioned domain-policy conditions. The architecture does not assume an oracle for semantic truth: missing or inapplicable support blocks authorization but does not prove that two referents are different.

\subsection{5.2 Constraint Binding and Its DIO/ODI Contribution}
The DIO/ODI Reference Model provides concepts, relationships, and diagnostic mappings. It does not automatically generate executable constraints. We propose a partial, domain-governed binding function

\begin{center}
\begin{adjustbox}{max width=\linewidth}
$\displaystyle \mu_{\Omega}:\mathcal C_{\mathrm{DIO/ODI}}\rightharpoonup\mathcal V_E,$
\end{adjustbox}
\end{center}
where \(\mathcal C_{\mathrm{DIO/ODI}}\) denotes the selected failure-mode distinctions in the restricted domain, and \(\mathcal V_E\) denotes the space of admissible domain-level constraint specifications for referent or entity class \(E\). Each element of \(\mathcal V_E\) is a specification that may contain one or more coordinated predicates; it is not assumed to be an atomic condition. The concrete collection configured for \(E\) in a given observation domain is denoted by \(V_E\subseteq\mathcal V_E\). The partial function \(\mu_\Omega\) maps a selected diagnostic distinction to one specification in \(\mathcal V_E\); the selected mappings inform the construction of \(V_E\) but do not automatically determine every predicate or policy condition required in an implementation. The mapping is a policy-design step, not automatic SKOS entailment. In this notation, \(E\) denotes the referent or entity class, \(V_E\) the configured collection of constraint specifications, and \(v_E\) the version counter for the evidence state—not a version of \(E\) or \(V_E\).

For this paper, \emph{reconstruction} means forming a context- and time-bound representation of a referent's identity from distributed representations and claims; it does not presuppose that one complete ground-truth profile is available. Digital Identity Optimization (DIO) is the broader framework for preserving and reconstructing coherent digital identity across such representations, while the Ontology of Digital Identity (ODI) organizes the concepts and relations used to describe that problem. In the present proposal, the domain of \(\mu_\Omega\) is restricted to the reconstruction-failure distinctions defined in Section 3 (RRF, SCF, TRF, and CAF) and the selected categories shown in the mapping below; it does not cover the whole DIO/ODI vocabulary.

General RDF graph validation can provide an implementation substrate for some domain-level constraints. SHACL specifies how RDF data graphs may be checked against declared shapes and conditions [30]. A suitable SHACL/SPARQL validator could therefore evaluate selected graph conditions, but SHACL does not itself supply the DIO/ODI diagnostic mapping, decide which identity proposition is required for a particular mutation, or establish the adequacy of the evidence-promotion process. The proposed \(\mu_\Omega\) is a design-time binding step, not a graph-validation engine or an automatic constraint-generation mechanism.

At minimum, a binding used for authorization should be:

\begin{itemize}
\item \textbf{Domain-scoped:} applicable to a declared entity class, observation context, and mutation class.
\item \textbf{Semantically typed:} explicit about which relation or evidence type it accepts; relatedness must not silently become identity equivalence.
\item \textbf{Temporally scoped where relevant:} continuity and state claims must be evaluated for the interval to which the mutation applies.
\item \textbf{Inspectable and versioned:} the selected concept-to-constraint mapping and the policy version used at authorization can be recorded and audited.
\item \textbf{Fail-closed for irreversible mutations:} insufficient authorization-eligible evidence yields \texttt{UNKNOWN\_EVIDENCE}, not implicit permission.
\end{itemize}
An illustrative binding is shown below. It is a domain-policy example, not a validated general-purpose mapping or a property automatically produced by the reference vocabulary. The diagnostic categories in the left column come from the DIO/ODI framing; the specific predicates and rules are supplied by the domain policy.

{\small
\begin{longtable}{@{}>{\raggedright\arraybackslash}p{0.23\dimexpr\linewidth-4\tabcolsep\relax} >{\raggedright\arraybackslash}p{0.34\dimexpr\linewidth-4\tabcolsep\relax} >{\raggedright\arraybackslash}p{0.43\dimexpr\linewidth-4\tabcolsep\relax}@{}}
\caption{Illustrative DIO/ODI diagnostic-to-constraint mapping for identity-changing mutations.}\\
\toprule
\textbf{DIO/ODI diagnostic distinction} & \textbf{Example domain-level constraint} & \textbf{Authorization consequence for an identity-changing mutation} \\
\midrule
\endfirsthead
\toprule
\textbf{DIO/ODI diagnostic distinction} & \textbf{Example domain-level constraint} & \textbf{Authorization consequence for an identity-changing mutation} \\
\midrule
\endhead
Entity Substitution / Merge Failure (RRF) & Lexical similarity or a shared commercial neighborhood does not by itself establish that two records denote the same referent. & Do not merge without claim-specific evidence establishing co-reference. \\
\addlinespace
Temporal Discontinuity / Misalignment (TRF) & A rename or continuity claim must apply to the relevant referent and validity interval. & Defer if temporal applicability or legal continuity is unresolved. \\
\addlinespace
Semantic Fragmentation (SCF) & Materially incompatible claims affecting identity are retained as a conflict until reconciled; clustering alone does not reconcile them. & Require reconciliation or defer the affected mutation. This mapping covers the identity-affecting subset; other attribute-level incompatibilities remain governed by domain policy. \\
\addlinespace
Contextual Displacement (CAF) & The same \texttt{brand\_of} relation may be adequate in a brand catalogue but insufficient for the legal-entity identity proposition required in a compliance context. & Preserve the typed relation; require evidence and constraints appropriate to the current observation purpose. \\
\addlinespace
Split Failure / valid continuity (RRF/TRF) & Documented, temporally applicable evidence may establish that distinct records refer to one continuing entity. & Permit resolution when the claim-specific evidence and all applicable constraints pass. \\
\addlinespace
\bottomrule
\end{longtable}
}
The proposed DIO/ODI-informed contribution is a \textbf{diagnostic-to-constraint mapping}: selected reconstruction-failure distinctions are used to specify which claim a mutation requires, which relations are insufficient for that claim, and when temporal or contextual conditions require deferral. This is a design proposal rather than an established advantage over generic authorization. Its incremental value depends on whether this mapping improves the safety/coverage trade-off relative to policies built from the same domain facts without the diagnostic structure.

\subsection{5.3 Evidence States and Promotion Contract}
Let \(P_{\mathrm{active}}\) be the evidence admitted for use by the Planner and \(P_{\mathrm{quar}}\) the evidence retained in quarantine outside that admitted set. \(P_{\mathrm{active}}\) may include records marked \texttt{UNVERIFIED}. Let \(q_m\) denote the identity or continuity proposition—or set of propositions—that must hold for mutation \(m\) to be authorized. Let \(P_{\mathrm{auth}}(q_m,\omega)\subseteq P_{\mathrm{active}}\) be the evidence eligible to bear on the claim-specific assessment of those propositions in observation domain \(\omega\), including its relevant temporal epoch or interval. This includes evidence that may support a proposition, establish a policy-relevant contradiction, or reveal an unresolved material conflict; the role taken by each item must be recorded in its claim-specific validation basis. Authorization eligibility is not itself a verdict and is not a global label asserting that a document is universally true. \texttt{UNVERIFIED} is an admission-level status: it means that the item has not yet been established as eligible for the proposition under assessment, not that it is universally unverified or false. Eligibility must be assessed relative to the particular proposition, observation domain, temporal applicability, and policy version. The same source may support one proposition while contradicting or being insufficient for another; use for a different proposition requires a separate policy assessment.

A retrieved observation first enters \texttt{QUARANTINE\_EVIDENCE}. Following basic admission checks, it may be added to \(P_{\mathrm{active}}\) with status \texttt{UNVERIFIED}, but that step does not promote it into \(P_{\mathrm{auth}}(q_m,\omega)\). Promotion requires a separate, auditable policy that assesses the identity or continuity claim needed for the proposed mutation.

\begin{figure}[htbp]
\centering
\begin{adjustbox}{max width=\linewidth}
\begin{tikzpicture}[
  main/.style={draw, rounded corners, align=center, font=\scriptsize, minimum height=9mm, inner sep=3pt},
  branch/.style={draw, rounded corners, align=center, font=\scriptsize, minimum height=10mm, inner sep=3pt},
  flow/.style={-{Stealth[length=1.8mm]}, line width=0.45pt},
  node distance=7mm and 6mm
]
\node[main, text width=1.35cm] (obs) {Retrieved observation};
\node[main, text width=1.9cm, right=of obs] (quar) {\texttt{QUARANTINE}\\\texttt{EVIDENCE}\\Admission conflict check};
\node[main, text width=1.65cm, right=of quar] (active) {$P_{\mathrm{active}}$\\\texttt{UNVERIFIED}};
\node[main, text width=2.05cm, right=of active, xshift=5mm] (promo) {Claim-specific role\\assessment and eligibility\\($q_m$, policy, provenance)};
\node[main, text width=1.85cm, right=of promo] (auth) {$P_{\mathrm{auth}}$\\$(q_m,\omega)$\\claim-specific eligible evidence};
\node[main, text width=2.15cm, right=of auth] (eval) {Commit-time evaluation\\graph constraints, scope, versions};

\draw[flow] (obs) -- (quar);
\draw[flow] (quar) -- (active);
\draw[flow] (active) -- (promo);
\draw[flow] (promo) -- (auth);
\draw[flow] (auth) -- (eval);

\node[branch, text width=2.25cm, below=11mm of quar] (qbranch) {\texttt{CONFLICT} or\\\texttt{INDETERMINATE}: retain in quarantine; adjudicate and re-check};
\node[branch, text width=2.25cm, below=11mm of active] (abranch) {Planner may use this material for reasoning; it is not positive authorization support};
\node[branch, text width=3.25cm, below=11mm of promo] (pbranch) {Support may satisfy positive requirements; eligible contradiction may establish \texttt{BREACH}; unresolved indications block \texttt{PASS} and yield \texttt{UNKNOWN\_EVIDENCE} unless \texttt{BREACH} takes precedence};
\node[branch, text width=2.6cm, below=11mm of eval] (ebranch) {\texttt{PASS} + valid scope/versions $\to$ commit\\\texttt{BREACH} $\to$ reject/escalate\\\texttt{UNKNOWN\_EVIDENCE} $\to$ defer};

\draw[flow] (quar.south) -- node[left, font=\tiny, align=right] {conflict /\\indeterminate} (qbranch.north);
\draw[flow] (active.south) -- (abranch.north);
\draw[flow] (promo.south) -- node[left, font=\tiny, align=right] {role\\assessed} (pbranch.north);
\draw[flow] (eval.south) -- (ebranch.north);
\end{tikzpicture}
\end{adjustbox}
\caption{Evidence admission and claim-specific eligibility assessment. The path from quarantine to $P_{\mathrm{active}}$ applies only when admission returns \texttt{NO\_CONFLICT\_DETECTED}; the path into $P_{\mathrm{auth}}$ applies only when a claim-specific assessment establishes eligibility in the recorded role. $P_{\mathrm{auth}}$ may contain evidence in support, contradiction, or unresolved-indication roles, each recorded for the proposition being assessed. Only support can satisfy a required positive condition; eligible contradiction evidence can establish a relevant conflict, while an unresolved indication blocks \texttt{PASS} and yields \texttt{UNKNOWN\_EVIDENCE} unless an established \texttt{BREACH} takes precedence under the status aggregation rule. Admission to $P_{\mathrm{active}}$ grants only \texttt{UNVERIFIED} status; it does not authorize a mutation. Commit remains conditional on graph constraints, authorized scope, and current versions.}
\label{fig:evidence-promotion-flow}
\end{figure}
The minimum promotion contract is:

\begin{enumerate}
\item It evaluates the \textbf{specific proposition} required for the candidate mutation, rather than document relevance, query match, or plausibility alone.
\item It records the source/provenance basis, the proposition assessed, the item's role (support, contradiction, or unresolved indication), temporal applicability where relevant, the validation basis, and the policy/validator version.
\item It does not treat Planner confidence, an LLM-generated \texttt{verified} label, or repetition across \texttt{UNVERIFIED} observations as sufficient authorization evidence.
\item It does not promote a claim solely because the Planner's hypothesis induced the query that retrieved supporting material. Corroboration used for promotion must have a claim-specific basis under the declared policy; simple agreement or document count is not enough by itself. Observations whose recorded provenance indicates a shared upstream query, hypothesis, source, or derivation path must not be counted as independent corroboration merely because they appear as separate items.
\item It must not infer positive support merely from the absence of a detected conflict.
\item The recorded provenance and validation basis for a promoted claim must not cite a shadow candidate or a claim whose own recorded basis cites shadow content. A promotion process must preserve enough recorded basis to apply this restriction transitively along that recorded chain.
\item Evidence becomes authorization-eligible only for the proposition, evidentiary role, and applicability established under the declared policy. Evidence that does not qualify as positive support cannot satisfy a required positive condition.
\end{enumerate}
These requirements constrain the promotion decision on the basis of recorded provenance; they do \textbf{not} claim that the system can reconstruct every causal dependency among retrieval actions and evidence origins. The shadow exclusion applies to the recorded validation/provenance chain used for promotion. Unrecorded or undisclosed derivations cannot be reliably screened by this contract and remain outside the guarantee; the broader dependency problem remains future research.

The validator implementation and its trust base are outside this paper's scope. This is a material assumption, not a solved subproblem. Eligibility metadata may be written or changed only by the designated promotion process or by a versioned policy/adjudication operation that triggers a new authorization decision; the Planner and raw tool outputs cannot directly assign or alter authorization eligibility. The architecture is conditional on the promotion process correctly applying the stated policy. It cannot protect against a validator that systematically misclassifies unsupported claims as authorized, a false authoritative source, or a policy that encodes the wrong identity conditions. In domains where the validity of a mention-to-identifier link is itself uncertain, the linkage must be treated as a claim requiring evidence; the existence of an identifier alone does not prove that the mention refers to its holder.

For evidence admission, a mutation-independent check uses the observation itself and the current evidence stores: \texttt{CheckAdmissionConflict(o, P\_active, P\_quar)} returns \texttt{CONFLICT}, \texttt{NO\_CONFLICT\_DETECTED}, or \texttt{INDETERMINATE}. \texttt{CONFLICT} denotes a supported material conflict; \texttt{INDETERMINATE} denotes a plausible material conflict that cannot be resolved on the available basis. \texttt{NO\_CONFLICT\_DETECTED} means only that this admission check did not establish a conflict. It is not positive evidence of identity, and the check cannot certify that all conflicts have been detected. The admission check does not take \texttt{q\_m} as an input because no mutation-specific proposition need be fixed at read time. A conflict or indeterminate result keeps the observation quarantined for adjudication; otherwise it may be admitted to \texttt{P\_active} as \texttt{UNVERIFIED}, without authorization promotion. An admission-level \texttt{CONFLICT} is a reason to quarantine the observation, not by itself a mutation-level \texttt{BREACH} verdict.

A quarantined item remains outside Planner-visible shared evidence until a policy-designated adjudication or versioned rule determines its disposition. Such a decision must be recorded, advance the evidence-state version, and trigger a fresh admission check before an item is released to \texttt{P\_active}; release confers only \texttt{UNVERIFIED} status, not claim-specific authorization eligibility. If no authorized reviewer or applicable rule is available, the item remains quarantined and may cause a mutation to be deferred. Escalation does not itself release evidence or authorize a commit.

At mutation evaluation, the supervisor performs a separate claim-specific check: \texttt{CheckClaimConflict(q\_m, P\_auth, P\_active, P\_quar)}. It returns \texttt{CLEAR} only when no unresolved material conflict bearing on the propositions in \texttt{q\_m} is detected. It returns \texttt{CONFLICT} only when claim-specific authorization-eligible evidence, assessed under the applicable policy, establishes a relevant contradiction or prohibitive condition. It returns \texttt{INDETERMINATE} when a possible material conflict is indicated only by \texttt{UNVERIFIED} evidence in \texttt{P\_active} or by evidence in \texttt{P\_quar}, or when conflicting authorization-eligible items cannot be reconciled under the policy’s authority and temporal rules. A detected \texttt{CONFLICT} or \texttt{INDETERMINATE} must prevent \texttt{PASS}. An \texttt{UNVERIFIED} item may therefore cause deferral by revealing an unresolved conflict, but it cannot itself establish a mutation-level \texttt{BREACH} or count as positive support for identity.

For action and graph evaluation, the statuses are:

\begin{itemize}
\item \texttt{PASS}: every required positive condition is supported by claim-specific authorization-eligible evidence, no applicable prohibition is established, and \texttt{CheckClaimConflict} returns \texttt{CLEAR}.
\item \texttt{BREACH}: authorization-eligible evidence establishes a relevant contradiction or violation of an explicit prohibitive constraint.
\item \texttt{UNKNOWN\_EVIDENCE}: required positive support is insufficient, or a material conflict remains unresolved or indeterminate.
\end{itemize}
The status aggregation rule is \textbf{\texttt{BREACH} takes precedence over \texttt{UNKNOWN\_EVIDENCE}, which takes precedence over \texttt{PASS}}: an established violation rejects the candidate even if other conditions are unknown; otherwise any missing required support or unresolved material conflict blocks a pass. Positive/sufficiency constraints state what must be supported to authorize a mutation. Prohibitive constraints state what must not be asserted or committed. A relation such as \texttt{brand\_of} is not, by itself, proof that two referents are different; it is insufficient to establish legal-entity identity unless a separate policy rule and evidence support that conclusion. \texttt{UNKNOWN\_EVIDENCE} blocks an irreversible mutation because authorization has not been established, not because the system has proven the identity claim false.

\subsection{5.4 Authorization Invariant and Assumptions}
Let \(m\) be a proposed mutation, \(q_m\) the proposition or set of propositions required to authorize it, \(G_{\mathrm{cand}}\) its candidate graph, \(\alpha_m\) the authorization record, and \(\sigma_c\) the consistent commit-time snapshot of the relevant graph, evidence stores, policy, shadow store if used, and their versions. The required proposition set is derived by the supervisor from the candidate operation, its target/effect scope, and the applicable versioned policy instance, which is configured for the relevant observation domain:

\begin{center}
\begin{adjustbox}{max width=\linewidth}
$\displaystyle q_m := \operatorname{DeriveRequiredClaims}(m,V_E,v_{\mathrm{pol}}).$
\end{adjustbox}
\end{center}
The resulting \(q_m\) is bound in \(\alpha_m\); the Planner cannot weaken or replace that derived set. The domain-specific policy instance represents the applicable observation context and temporal conditions. Accordingly, \(\omega\) in \(P_{\mathrm{auth}}(q_m,\omega)\) is fixed by the policy instance bound through \(v_{\mathrm{pol}}(\alpha_m)\); it is not an unbound runtime input. The claim-specific evidence assessment evaluates applicability to \(q_m\) within that bound context. This is the operational bridge from domain-configured constraints to the specific propositions that must be supported for a mutation. The ideal set \(\operatorname{Adm}(E,\omega)\) defined in Section 5.1 is not evaluated directly at commit time. Rather, the design-time mapping \(\mu_\Omega\) selects explicit constraints from the DIO/ODI diagnostic vocabulary, the versioned policy instantiates the required propositions and constraints, and the commit contract checks their satisfaction against claim-specific authorization-eligible evidence. Consequently, a \texttt{PASS} result establishes compliance with the configured contract; it does not prove membership in the ideal admissible set or the semantic truth of the underlying evidence. An operation is treated as a protected identity-changing mutation whenever its realized or potential shared effects can alter the graph's identity interpretation. This applies to merges, splits, and corrective un-merges or other reversals when they alter shared identity interpretation; the versioned domain policy defines the required propositions separately for each mutation type. Reversing a prior merge does not, by itself, establish that the affected referents are distinct. If the effect class is ambiguous, it is treated as protected rather than classified as harmless or reversible by the Planner.

The authorization record binds the domain-policy version \(v_{\mathrm{pol}}(\alpha_m)\). The functions \(\operatorname{ScopeValid}\) and \(\operatorname{VersionsValid}\) are defined in Section 6. \textbf{Under Assumptions 1--7 below, the proposed commit invariant is:}

\begin{center}
\begin{adjustbox}{max width=\linewidth}
$\displaystyle \begin{aligned}
\operatorname{Commit}(m)\Rightarrow{}&
\operatorname{EvaluateGraphConstraints}(V_E^{(v_{\mathrm{pol}}(\alpha_m))},q_m,
P_{\mathrm{auth}}^{(\sigma_c)}(q_m,\omega),P_{\mathrm{active}}^{(\sigma_c)},P_{\mathrm{quar}}^{(\sigma_c)},G_{\mathrm{cand}}^{(\sigma_c)})=\texttt{PASS}\\
&\land\operatorname{ScopeValid}(m,\alpha_m)\land\operatorname{VersionsValid}(\alpha_m,\sigma_c).
\end{aligned}$
\end{adjustbox}
\end{center}
\texttt{EvaluateGraphConstraints} evaluates both the candidate graph against applicable constraints and the evidence-to-claim support needed to authorize the proposed mutation; it is not merely a graph-shape validator. Each required proposition in \(q_m\) must be covered separately by at least one item in \(P_{\mathrm{auth}}(q_m,\omega)\) whose recorded claim-support and validation basis is applicable to that proposition, including temporal applicability where required. The authorization record \(\alpha_m\) must bind \(q_m\) and, for each required proposition, identify the promotion/validation record or stable record identifier relied on for its claimed support. Mere membership in \(P_{\mathrm{auth}}\) without proposition-specific support is insufficient. \texttt{EvaluateGraphConstraints} returns \texttt{PASS} only if all required positive conditions are supported by claim-specific authorization-eligible evidence, no applicable prohibition is established, and \texttt{CheckClaimConflict} returns \texttt{CLEAR} for the evidence stores in the same snapshot \(\sigma_c\). It returns \texttt{BREACH} for an established policy violation and \texttt{UNKNOWN\_EVIDENCE} for insufficient positive support or a conflict reported as unresolved or indeterminate. The check is not assumed to detect contradictions absent from, or unrecognized in, its inputs. A conforming implementation must reject or defer unless the result is \texttt{PASS}, the realized operation falls within its authorized scope, and all relevant versions remain current through the commit point. \texttt{VersionsValid} must verify both that the authorization record was based on the relevant snapshot and that the graph, evidence-state, policy, and (if used) shadow versions still match that snapshot atomically at commit; validation against a stale snapshot alone is insufficient. This is an \textbf{interface invariant}: it specifies the protected commit path, not a theorem that the evidence or policy is true.

The conditional property relies on seven stated conditions with different logical roles. Assumptions 4–6 below are the enforcement and binding premises of the commit implication. Assumptions 1, 2, and 7 are adequacy conditions for interpreting a \texttt{PASS} as support for an identity claim; they do not independently establish semantic correctness. The fail-closed behavior in Assumption 3 restates the \texttt{PASS}/\texttt{UNKNOWN\_EVIDENCE} outcome contract: insufficient support or unresolved material conflict cannot authorize an irreversible mutation.

\begin{enumerate}
\item \textbf{Domain-policy adequacy:} \(V_E\) expresses relevant identity-preserving constraints for the entity class, context, time, and mutation type.
\item \textbf{Claim-specific promotion:} evidence enters \(P_{\mathrm{auth}}(q_m,\omega)\) only under the promotion contract in Section 5.3, including its limits on Planner/query-conditioned promotion and corroboration by \texttt{UNVERIFIED} material.
\item \textbf{Fail-closed authorization:} insufficient evidence or an unresolved material conflict does not authorize an irreversible identity change.
\item \textbf{Enforced and conservative mutation classification:} every relevant production write passes through the supervisor, and any operation that may alter shared identity interpretation is treated as protected. Unmanaged write paths are outside the guarantee.
\item \textbf{Derived authorization propositions:} \(q_m\) is derived by the supervisor from the mutation, its target/effect scope, and the versioned domain policy; the Planner cannot supply a substitute claim set or weaken the derived authorization obligation.
\item \textbf{Scope and version integrity:} the operation, target, domain-policy version, graph state, complete evidence state (including quarantine), authorization eligibility, and shadow state if used remain valid through the commit point.
\item \textbf{Recorded-basis and eligibility integrity:} the promotion process controls eligibility metadata and applies the recorded-provenance restrictions in Section 5.3; the guarantee does not extend to derivations absent from that recorded basis.
\end{enumerate}
The property is conditional on the promotion process applying the stated contract. It cannot protect against an invalid authoritative source, a validator that misclassifies unsupported claims, or a policy that encodes the wrong identity conditions. It does not guarantee that upstream reasoning is uncontaminated or that every material conflict is detected.

\subsection{5.5 Boundary and Future Direction: Dependency-Aware Evidence Flow}
A broader design could also control how evidence and intermediate hypotheses influence subsequent actions. That is a different problem from authorization at the commit boundary. A plausible future research direction is a dependency-aware information-flow layer that investigates three requirements:

\begin{enumerate}
\item \textbf{Derivation lineage:} retain links among observations, the actions or queries that produced them, the hypotheses that conditioned those actions, and claims subsequently derived from them.
\item \textbf{Dependency-aware corroboration:} avoid treating several observations as independent confirmation solely because they appear in separate documents or endpoints when their derivation paths are materially dependent.
\item \textbf{Controlled feedback to the Planner:} preserve or expose dependency status when evidence is reused for planning, and test when dependent evidence should be withheld, quarantined, discounted, or sent through an alternative validation path.
\end{enumerate}
This is a research agenda, not a mechanism implemented or validated here. It would require a clear dependency model, an explicit information-flow contract, and evaluation using controlled hypothesis-conditioned perturbations or trajectory replay. Its effectiveness should not be implied by the commit-boundary authorization invariant.

\section{6. Abstract Interface Specification}
We propose the \textbf{MVDIO Supervisor Layer} as a governance gateway for Agentic ER [17]. The architecture has a pre-execution authorization phase and a post-execution validation phase. Its purpose is to govern evidence admission and persistent mutation authorization; it is not a controller for the agent's complete reasoning trajectory.

\subsection{6.1 State, Scope, and Version Contracts}
The supervisor distinguishes four action classes:

\begin{itemize}
\item \texttt{READ}: retrieval or exploration within a declared scope.
\item \texttt{STAGE\_LOCAL}: local preparation with no shared-state effects.
\item \texttt{REVERSIBLE\_WRITE}: a provisional operation authorized only for an isolated staging target, such as \texttt{G\_shadow}, where effects are separated from production identity state, Planner-visible shared identity state, downstream consumers, and all \texttt{P\_auth} inputs, and can be safely discarded. It does not authorize publication to the production graph.
\item \texttt{IRREVERSIBLE\_WRITE}: the protected operational class for any operation whose realized or potential shared effects can persistently alter identity interpretation, such as merging legal-entity nodes or publishing an identity-changing edge. The label means that authorization is required at the shared persistence boundary; it does not assert that technical rollback is impossible. A candidate may be classified as \texttt{REVERSIBLE\_WRITE} only when the isolation conditions are positively established under policy; if classification or isolation is uncertain, the operation is treated as \texttt{IRREVERSIBLE\_WRITE}.
\end{itemize}
A \texttt{READ} result must not directly overwrite Planner-visible shared memory. The raw result first enters \texttt{QUARANTINE\_EVIDENCE}; after admission checks, it may be admitted to \(P_{\mathrm{active}}\) as \texttt{UNVERIFIED}. The Planner may then use it for reasoning. This design does not claim to prevent the Planner from being influenced by it. Only the separate promotion contract determines whether it is eligible for \(P_{\mathrm{auth}}(q_m,\omega)\). \texttt{UNVERIFIED} is an admission-level status, not a universal judgment about an item's truth: for a proposition under assessment, it means that claim-specific eligibility has not been established. Admission as \texttt{UNVERIFIED} is never itself promotion; eligibility is specific to the proposition, observation domain, temporal applicability, and policy version, and remains subject to the policy/version checks.

The authorization record binds the permitted operation type, target and effect scope, and applicable domain-policy version \(v_{\mathrm{pol}}\). For protected graph mutations it also binds the supervisor-derived required-claim set \(q_m\), graph version \(v_G\), evidence-state version \(v_E\), and shadow version \(v_S\) when a shadow layer is used. Version bindings are action-specific: a READ authorization does not bind a pre-retrieval \(v_E\) or \(v_G\) whose change would force the retrieval to be replayed; its returned observation is admitted through the current-state protocol in Algorithm 2. A STAGE\_LOCAL authorization binds shared-state versions only when the declared local operation explicitly depends on them. A REVERSIBLE\_WRITE binds \(v_S\), the policy version \(v_{\mathrm{pol}}\), and the versions of any separately versioned isolation state relevant to staging. Use one evidence-state version, \(v_E\), for the complete relevant evidence state: it advances whenever \(P_{\mathrm{active}}\) or \(P_{\mathrm{quar}}\) changes, including quarantine intake, removal, or adjudication outcome, and whenever authorization eligibility is promoted, demoted, or revoked. Every such change invalidates protected-write authorization based on the prior \(v_E\). \texttt{ScopeValid} checks the realized operation type, target, and allowed effects against the record. \texttt{VersionsValid} checks all versions bound and relevant to that operation at the required linearization or commit point. The policy is versioned and checked at authorization and at any later decision that relies on it.

We define an abstract \texttt{AtomicSnapshot} operation that returns a consistent logical snapshot of the production graph, both evidence stores (\texttt{P\_active} and \texttt{P\_quar}), the authorization-eligibility state, the applicable policy, the staging-isolation state when a \texttt{REVERSIBLE\_WRITE} is authorized, the optional shadow store, and the versions of these state components. A conforming implementation may use different underlying mechanisms but must satisfy the declared snapshot semantics. A production commit must atomically validate relevant versions and apply the candidate against the same commit point. Merely checking versions at separate earlier times is insufficient.

A stable \texttt{operation\_id} supports idempotency and reconciliation after uncertain network outcomes; it does not by itself provide atomicity. The transaction layer reports one of four outcomes: \texttt{COMMITTED}, \texttt{VERSION\_CONFLICT}, \texttt{COMMIT\_FAILED}, or \texttt{TIMEOUT\_UNKNOWN}. These are transaction outcomes, not supervisor directives or semantic identity verdicts. \texttt{COMMITTED} maps to the supervisor directive \texttt{COMMITTED}; a version or snapshot conflict triggers \texttt{RETRY\_REEVALUATE} against a fresh snapshot, subject to one common retry budget across all retry triggers, and exhaustion yields \texttt{TRANSACTION\_RETRY\_EXHAUSTED}; a confirmed \texttt{COMMIT\_FAILED} with no committed effect maps to \texttt{SYSTEM\_FAILURE\_ESCALATE}; and \texttt{TIMEOUT\_UNKNOWN} maps to \texttt{COMMIT\_UNKNOWN\_RECONCILE} until the outcome can be reconciled. Any human adjudication that changes the case disposition must result in a new authorization decision against current versions; approval must not silently revive a stale candidate. A confirmed infrastructure failure yields \texttt{SYSTEM\_FAILURE\_ESCALATE}, not a semantic identity verdict.

\subsection{6.2 Phase 1 — Pre-Execution Authorization}
The abstract pre-execution interface evaluates the proposed action against its declared scope, the domain policy, and the evidence eligible for the relevant authorization proposition. For a persistent identity-changing operation, it returns an authorization record only after the applicable action constraints pass.

\begin{Verbatim}[fontsize=\footnotesize,breaklines=true,breakanywhere=true]
Algorithm 1: Abstract Pre-Execution Authorization
Input: action a_t, domain policy V_E and version v_pol,
       P_active, P_quar, current graph/evidence/policy versions, any separately
       versioned staging-isolation state when relevant, and v_S if used
Output: directive and, if authorized, a bound authorization record

1. If the operation type is unsupported, return REJECT_UNSUPPORTED_OPERATION.
2. If the action has unmanaged external effects, return OUT_OF_BAND_AUTHORIZATION_REQUIRED;
   do not execute that effect under this interface.
3. Classify the operation from its schema, potential effects, and applicable
   policy. An operation that may alter shared identity interpretation is
   IRREVERSIBLE_WRITE. Classify as REVERSIBLE_WRITE only when isolated staging
   and safe discard are positively established under the policy; ambiguity is protected.
4. For READ or STAGE_LOCAL, authorize only the declared scope; do not treat this
   as validation of returned content. Return the applicable directive (READ_AUTHORIZED
   for READ; STAGE_LOCAL_AUTHORIZED for STAGE_LOCAL) together with the corresponding
   scope-bound authorization record, including the operation type, authorized scope,
   and versions required for that action class.
5. For REVERSIBLE_WRITE, authorize staging only to G_shadow after verifying
   isolation from production identity state, Planner-visible shared identity
   state, downstream consumers, and P_auth inputs. Bind the permitted type, target/effect
   scope, policy, and relevant state versions in a stage-only authorization record; it
   cannot authorize production commit. Return that record. If isolation cannot be established, do not stage;
   classify the intended shared effect as IRREVERSIBLE_WRITE and continue at Step 6
   through the protected authorization path before execution.
6. For a protected graph mutation, derive q_m from the mutation, target/effect
   scope, and versioned policy. Bind this derived set in the authorization record;
   do not substitute a Planner-supplied claim set.
7. Perform an action-level preflight against the current graph/evidence/policy snapshot:
   assess claim-specific evidence bearing on q_m, applicable constraints that can be
   checked before execution, declared scope, and claim-specific conflicts. Do not
   evaluate a not-yet-realized candidate graph; Algorithm 2 performs the full graph
   evaluation after constructing the isolated candidate.
8. If evaluation is BREACH, reject the candidate and return
   REJECT_CANDIDATE_AND_ESCALATE.
9. If evaluation is UNKNOWN_EVIDENCE, do not issue authorization. Return
   HALT_HITL only when policy requires review and a reviewer is available;
   otherwise return DEFER_UNRESOLVED. No protected candidate is staged or retained
   in G_shadow through this branch.
10. On PASS, bind target, operation/effect scope, q_m, policy version, and the
    relevant graph, evidence-state, and shadow versions in the authorization record.
    For every required proposition in q_m, also bind the stable identifier of the
    promotion/validation record whose claim-specific support was relied on.
11. Return the authorization record for post-execution validation.
\end{Verbatim}
The pseudocode is an abstract contract, not a production implementation recipe. It intentionally omits implementation-specific source-ranking rules, thresholds, proprietary prompts, and validator internals. Human review may be part of the domain policy for difficult cases; the paper does not assume that a reviewer is always available or that human judgment is infallible [14, 17]. A human may supply or assess evidence, promote a claim under the declared policy, or authorize a versioned policy change; a reviewer does not silently override \texttt{UNKNOWN\_EVIDENCE} or bypass the commit invariant. Any such action requires a new authorization decision against current versions. When review is unavailable, the case remains deferred rather than being silently approved.

\subsection{6.3 Phase 2 — Post-Execution Validation and Commit}
The post-execution interface validates the realized outcome against the authorization record. \texttt{MatchesAuthorizedScope} checks action type, target, and permitted effects; it is not a content-truth test. For candidate graph writes, the supervisor constructs an isolated candidate, re-evaluates the candidate graph under the same applicable policy and claim-specific authorization evidence, and verifies that the policy/evidence/graph versions have not changed in a way that invalidates the authorization. A policy-version change requires re-evaluation just as a graph- or evidence-version change does.

\begin{Verbatim}[fontsize=\footnotesize,breaklines=true,breakanywhere=true]
Algorithm 2: Abstract Post-Execution Validation
Input: realized operation op_t (including returned observation o_{t+1} when op_t is READ),
       authorization record (including bound q_m for protected graph mutations),
       production graph G_prod, P_active, P_quar,
       domain policy V_E, optional shadow store G_shadow, retry_count, retry_budget,
       operation_id
Output: post-execution directive

The protected graph-write path applies only after the READ, STAGE_LOCAL, and
REVERSIBLE_WRITE branches return. A REVERSIBLE_WRITE may stage only to an
isolated G_shadow and never authorizes publication to G_prod.

1. If op_t does not exactly match the authorized action/effect scope,
   stop processing the result or candidate and return REJECT_CANDIDATE_AND_ESCALATE.
2. Obtain a consistent snapshot sigma_c of the relevant production graph,
   P_active, P_quar, authorization-eligibility state, policy, staging-isolation
   state when a REVERSIBLE_WRITE is evaluated, optional G_shadow, and their
   versions, including v_E for all active/quarantined evidence changes.
3. For REVERSIBLE_WRITE or a protected graph write, if any version bound to the
   authorization record is stale or incompatible with sigma_c, increment retry_count.
   If retry_count reaches retry_budget, return TRANSACTION_RETRY_EXHAUSTED; otherwise
   return RETRY_REEVALUATE and restart Algorithm 1 against a fresh snapshot with a new
   authorization record. Do not replay a stale candidate. For READ, if its policy/scope
   authorization is no longer valid, reject the result and return REJECT_CANDIDATE_AND_ESCALATE.
   Otherwise do not reject it merely because unrelated shared evidence or graph versions
   changed during retrieval; Step 4 linearizes admission against the current evidence state
   and must not re-execute the READ. For STAGE_LOCAL, if an explicitly bound local
   dependency version is stale, discard the local artifact, increment retry_count, and
   return TRANSACTION_RETRY_EXHAUSTED when the budget is used or RETRY_REEVALUATE
   against a fresh snapshot otherwise. If no relevant local dependency is stale, proceed
   to Step 5; shared-state versions are not required unless the operation depends on them.
4. If the action was READ:
   a. atomically place the returned material in QUARANTINE_EVIDENCE and advance v_E;
   b. run CheckAdmissionConflict(o_{t+1}, P_active, P_quar) against a consistent snapshot;
      this check does not depend on a mutation-specific q_m;
   c. on CONFLICT or INDETERMINATE, retain the item in quarantine and return
      QUARANTINE_ONLY, provided the evidence-state version remains the one checked;
      if it changed, increment retry_count and repeat the admission check against a
      fresh snapshot, or return TRANSACTION_RETRY_EXHAUSTED when the budget is used;
   d. if no conflict is detected, use compare-and-swap on that checked evidence-state
      version to atomically remove the item from P_quar, admit it to P_active as
      UNVERIFIED, and advance v_E; if the compare-and-swap fails, increment retry_count
      and repeat the check against a fresh snapshot, or return TRANSACTION_RETRY_EXHAUSTED
      when the budget is used;
   e. do not promote it into P_auth as part of this admission step; return
      ADMIT_EVIDENCE_COMMITTED.
5. If the action was STAGE_LOCAL, confirm that no shared-state effects occurred.
   If confirmed, return STAGE_LOCAL_CONFIRMED; otherwise return
   REJECT_CANDIDATE_AND_ESCALATE.
6. If op_t is a REVERSIBLE_WRITE, verify that its authorization permits staging
   to G_shadow and that the realized effects remain isolated from production,
   Planner-visible shared identity state, downstream consumers, and P_auth inputs.
   Revalidate the authorized target/effect scope and isolation conditions. Perform
   the version comparison and shadow write as one atomic operation: apply only to
   G_shadow if v_S, v_pol, and every other staging-relevant version bound to the
   authorization record—including the version of any separately represented
   isolation state—still match sigma_c at the operation's linearization point.
   On success, advance v_S and return SHADOW_STAGED_PROVISIONAL. If any bound
   version no longer matches, do not stage; increment retry_count and, if the retry
   budget is not exhausted, return RETRY_REEVALUATE and restart Algorithm 1 against
   a fresh snapshot. If the retry budget is exhausted, return
   TRANSACTION_RETRY_EXHAUSTED. If the authorized scope or isolation conditions
   cannot be verified, discard the candidate and return
   REJECT_CANDIDATE_AND_ESCALATE; any shared-identity attempt must restart through
   Algorithm 1 as an IRREVERSIBLE_WRITE.
7. For a protected graph write, apply op_t only to an isolated candidate graph G_cand.
8. Re-derive the required claim set from op_t and the policy in sigma_c. If it
   differs from the q_m bound in the authorization record, discard the candidate
   and return REJECT_CANDIDATE_AND_ESCALATE.
9. Evaluate G_cand against the policy and evidence state in the same sigma_c,
   using the bound q_m, P_auth(q_m, omega), P_active, P_quar, and the
   claim-specific conflict check. For each required proposition, verify that the
   promotion/validation record identifier bound in alpha_m is present in the same
   snapshot and that its claim-specific support, role, applicability, and policy
   version still hold in P_auth. Re-derive neither support nor provenance from a
   later snapshot. Map CheckClaimConflict=CONFLICT to BREACH and
   CheckClaimConflict=INDETERMINATE to UNKNOWN_EVIDENCE; either result blocks PASS.
10. If the result is BREACH, discard the candidate and return
    REJECT_CANDIDATE_AND_ESCALATE.
11. If the result is UNKNOWN_EVIDENCE, do not commit or stage the protected candidate
    in G_shadow; defer the identity-changing decision. Return HALT_HITL only when
    policy requires adjudication and a reviewer is available; otherwise return
    DEFER_UNRESOLVED.
12. If the result is PASS, atomically attempt commit only if the authorized scope
    remains valid and the current graph version v_G, evidence-state version v_E,
    policy version v_pol, and shadow version v_S (if used and bound in the
    authorization record) still match sigma_c at the commit point. On successful commit,
    advance v_G and return COMMITTED. If atomic version validation fails, handle it
    as VERSION_CONFLICT under Step 13. A confirmed COMMIT_FAILED with no committed
    effect is handled under Step 13; an uncertain outcome is handled as TIMEOUT_UNKNOWN.
13. On VERSION_CONFLICT, increment retry_count; if the retry budget is exhausted,
    return TRANSACTION_RETRY_EXHAUSTED, otherwise return RETRY_REEVALUATE and
    restart Algorithm 1 against a fresh snapshot with a new authorization record.
    This same retry budget applies to retry triggers in Steps 3 and 6 and to version changes
    or admission compare-and-swap failures in Step 4; for Step 4, rerun the admission
    check against a fresh snapshot rather than re-executing the READ action. On TIMEOUT_UNKNOWN, return
    COMMIT_UNKNOWN_RECONCILE and freeze until the operation_id outcome is reconciled.
    On confirmed COMMIT_FAILED, return SYSTEM_FAILURE_ESCALATE; do not treat it as
    a semantic identity verdict.
\end{Verbatim}
\textbf{Lifecycle semantics.} Pre-execution directives, post-execution outcomes, and case statuses are distinct; the transaction-outcome mapping is defined in Section 6.1. \texttt{READ\_AUTHORIZED} and \texttt{STAGE\_LOCAL\_AUTHORIZED} are pre-execution scope decisions, not validation of content or completed operations. \texttt{STAGE\_LOCAL\_CONFIRMED} means the local operation completed without shared-state effects. \texttt{REJECT\_UNSUPPORTED\_OPERATION} denotes rejection of an unsupported action type; \texttt{OUT\_OF\_BAND\_AUTHORIZATION\_REQUIRED} means execution is withheld pending authorization outside this interface. \texttt{COMMITTED} means that the protected candidate passed commit-time validation and was committed within the authorized scope, with the graph version advanced; it is not a claim of semantic truth beyond the stated policy and evidence assumptions. \texttt{QUARANTINE\_ONLY} keeps an observation outside Planner-visible shared evidence pending adjudication; it does not label the item false or malicious. \texttt{ADMIT\_EVIDENCE\_COMMITTED} means an item entered \(P_{\mathrm{active}}\) as \texttt{UNVERIFIED}, not that it became authorization-eligible. \texttt{SHADOW\_STAGED\_PROVISIONAL} means a candidate was staged only in isolated \(G_{\mathrm{shadow}}\); it is neither a production commit nor permission to publish and any later production mutation requires a new protected authorization decision. \texttt{DEFER\_UNRESOLVED} terminates the current automated attempt but leaves the case available for later review or explicitly authorized retry. \texttt{HALT\_HITL} is emitted only when the policy requires human adjudication and a reviewer is available; if no reviewer is available, return \texttt{DEFER\_UNRESOLVED}. A human decision may update evidence eligibility or the versioned policy but cannot bypass the invariant; it requires a new authorization decision. \texttt{RETRY\_REEVALUATE} requires a fresh snapshot and a new authorization decision. \texttt{TRANSACTION\_RETRY\_EXHAUSTED} is returned when the common retry budget is exhausted, regardless of which version/snapshot check triggered it; \texttt{COMMIT\_UNKNOWN\_RECONCILE} freezes the case until a timeout outcome is reconciled; and \texttt{SYSTEM\_FAILURE\_ESCALATE} reports a confirmed infrastructure failure. These are not semantic verdicts. \texttt{REJECT\_CANDIDATE\_AND\_ESCALATE} discards a not-yet-committed candidate; it does not imply rollback of a previously committed mutation. Its destination is a policy-defined adjudication or operations queue; escalation does not itself change evidence eligibility or authorize a retry. \texttt{SHADOW\_STAGED\_PROVISIONAL} applies only to an explicitly authorized \texttt{REVERSIBLE\_WRITE} under the Section 6.1 isolation conditions, not to a protected candidate deferred as \texttt{UNKNOWN\_EVIDENCE}.

The claimed property is limited to the protected persistence boundary: \texttt{UNVERIFIED} evidence is not eligible to satisfy positive authorization conditions unless and until it is separately promoted for the specific claim under the stated contract. Evidence may still influence the Planner, and the interface does not establish that every bad claim or material conflict will be recognized, that provenance assertions are truthful, or that the agent's internal state remains uncontaminated.

\section{7. Illustrative Case: Corporate Entity Resolution in B2B Data Lakes}
This hypothetical scenario concerns Enterprise Knowledge Graph construction for risk and compliance [4, 12, 19]. It illustrates the distinction between an identity relation and other relations. It is not an observation from a deployed system.

\subsection{7.1 Scenario and Domain Constraints}
An Agentic ER system is resolving the legal entity \textbf{Acme Global}, which has acquired \textbf{BetaTech}. A separate startup, \textbf{Acme AI, Inc.}, operates in a related technical sector. \textbf{Acme Cloud} is a brand of Acme Global (\texttt{Acme Cloud brand\_of Acme Global}). Distributed evidence includes legal records, CRM entries, and noisy media reports that conflate Acme Cloud with Acme AI, Inc. because of lexical overlap.

For a financial-compliance observation domain, the policy configures explicit identity-preserving constraints \(V_E\) [6, 26, 27]. The following are illustrative policy-local predicates for this case, not claims that these exact relation labels are canonical predicates in the DIO/ODI vocabulary. Relevant relation types include:

\begin{itemize}
\item \texttt{same\_legal\_entity}: an identity-equivalence relation subject to claim-specific authorization;
\item \texttt{renamed\_to}: continuity of the same legal entity over time, subject to temporal validity;
\item \texttt{brand\_of}: a brand-to-legal-entity relation that does not authorize identity merge;
\item \texttt{absorbed\_by}: an acquisition relation that does not by itself establish \texttt{same\_legal\_entity} or legal continuity. It is a non-sufficient relation, not by itself a prohibitive finding that the identity claim is false; absent sufficient continuity evidence the result is \texttt{UNKNOWN\_EVIDENCE}, while \texttt{BREACH} requires separate authorized evidence or an explicit policy condition establishing a contradiction.
\end{itemize}
\textbf{Legal Lineage Constraint.} Distinct representations may be merged through an \texttt{IRREVERSIBLE\_WRITE} only when authorization-eligible evidence establishes the required \texttt{same\_legal\_entity} condition or applicable legal continuity through \texttt{renamed\_to}. A temporally valid Legal Entity Identifier (LEI) can provide relevant legal-entity evidence [26, 27], but the existence of an LEI does not by itself prove that a particular mention belongs to its holder. A brand relation or acquisition edge alone does not establish that two entities are identical.

\subsection{7.2 Failure Case: Relatedness Is Mistaken for Identity}
An unconstrained agent identifies lexical or semantic similarity between Acme Cloud and Acme AI, Inc. A retrieval step returns a speculative blog post discussing a commercial partnership. A naive matcher treats evidence of a partnership involving the Acme Cloud brand as evidence of referential identity and concludes that Acme AI, Inc. is identical to Acme Cloud. It then propagates the mistaken equivalence to Acme Global. This is an illustrative Merge Failure and Contextual Displacement under the DIO/ODI distinctions [7, 18].

For this example, assume that \(P_{\mathrm{auth}}\) contains no claim-specific evidence establishing \texttt{same\_legal\_entity} or an applicable \texttt{renamed\_to} continuity claim between the two candidate referents. The speculative partnership post is available only as \texttt{UNVERIFIED} context, and no authorization-eligible evidence independently establishes the required identity proposition. The brand and partnership relations therefore cannot satisfy the positive condition; the graph evaluation returns \texttt{UNKNOWN\_EVIDENCE}, and the irreversible merge is deferred. If claim-specific authorized evidence instead established that the two legal entities were distinct, the result would be \texttt{BREACH}. This is an illustration of the commit-boundary property, not evidence that all variants of apparent verification will be intercepted.

\textbf{Hypothesis-conditioned accumulation.} Suppose the same mistaken hypothesis also prompts searches of a trade directory and a news index. Three returned items appear to reinforce the proposed merge, but their recorded provenance links lead back to the same promotional announcement and none independently establishes \texttt{same\_legal\_entity} or valid legal continuity. The items may be admitted as \texttt{UNVERIFIED} observations, but their number and apparent agreement do not make them three independent grounds for promotion. Under promotion-contract items 3–4, they remain ineligible as positive identity support unless the declared policy accepts a separately recorded, claim-specific basis. The case therefore remains \texttt{UNKNOWN\_EVIDENCE} and is deferred. This example tests the recorded-provenance restriction in the promotion contract; if the relevant derivation links are absent from the record, the architecture does not guarantee that the dependency will be detected.

\subsection{7.3 Temporal Scope Case: Reused Source-Local Identifier}
Consider a second hypothetical case involving a source-local supplier key, \texttt{SUP-041}. A source's change log shows that the key was assigned to Supplier A during one validity interval, retired, and later reassigned to unrelated Supplier B. A flat rule that treats equality of this source-local identifier as sufficient identity evidence would merge the records despite their non-overlapping assignments. A TRF-informed constraint treats the identifier as temporally scoped: equality outside the same valid assignment interval cannot satisfy the identity proposition. If the reassignment record is authorization-eligible and establishes different assignments, evaluation returns \texttt{BREACH}; if the interval or assignment remains unresolved, it returns \texttt{UNKNOWN\_EVIDENCE}. Neither outcome authorizes the merge.

This example distinguishes a temporally typed constraint from a \textbf{flat identifier-equality rule}, not from a competent baseline that already encodes the same validity intervals. It illustrates how a DIO/ODI diagnostic can make an identity-preserving condition explicit; it does not demonstrate empirical superiority over well-designed temporal entity resolution.

\subsection{7.4 Contrast Case: Documented Identity Continuity}
Consider a paired hypothetical case in which two records have different names because the same legal entity was formally renamed. If appropriate legal records establish the continuity claim for the relevant time interval, and claim-specific evidence satisfies the configured constraints, the guardrail may return \texttt{PASS} and permit the authorized resolution. If the available record merely establishes a commercial acquisition or brand relationship, that is insufficient. The contrast illustrates that the policy is not a blanket rule rejecting every uncertain match; it is a typed mapping between the relationship asserted, the evidence available, and the mutation requested.

\section{8. Proposed Evaluation Protocol}
This section proposes an evaluation plan for future empirical validation. No experimental results are claimed. The central question is whether DIO/ODI-informed mutation constraints improve the safety/coverage trade-off beyond generic authorization controls and conventional identifier rules, under comparable resources.

\subsection{8.1 Baselines and Corpus Construction}
The evaluation should compare at least four conditions:

\begin{itemize}
\item \textbf{Baseline A — Unconstrained Agentic ER:} a documented agent using retrieval tools and an ER matcher without the proposed mutation-authorization boundary.
\item \textbf{Baseline B — Conventional identifier/default-deny rules:} the same task with standard identifier constraints and abstention when required evidence is missing.
\item \textbf{Baseline C — Generic authorization boundary:} the same two-phase mutation gate and claim-specific evidence separation, but with conventional policy rules authored without the DIO/ODI diagnostic mapping.
\item \textbf{Challenger — DIO/ODI-informed guardrail:} the same gate, evidence-promotion process, validator class and configuration, domain facts, and resource budget as C, with those same prespecified domain facts organized through the proposed mapping from referential, semantic, temporal, and contextual reconstruction-failure distinctions to mutation-specific constraints.
\end{itemize}
Baseline C and the Challenger must use the same case inventory, underlying domain-fact annotations, evidence-promotion process, validator class and configuration, action resources, and decision budget. Any non-policy component or configuration that cannot be shared must be declared in advance and recorded as a prespecified difference. Both conditions use the same validator class and configuration; the mutation-specific proposition sets \(q_m\) may differ because policy organization is the intended treatment pathway, and the resulting differences in required propositions, their derivation, and coverage must be reported rather than hidden under the phrase “held constant.” The comparison holds the validator process constant, not the policy outputs that it evaluates.

The case inventory and its diagnostic and taxonomy-neutral strata must be constructed or independently reviewed by a party not involved in authoring either compared policy. Construction and review must take place without access to frozen condition policies. At least one taxonomy-neutral stratum, defined from observable case properties without DIO/ODI labels, should be reported alongside the prespecified diagnostic strata. The diagnostic strata may include temporally reused source-local identifiers, context-dependent relation sufficiency, and hypothesis-conditioned observations with recorded shared upstream origins. The party constructing or adjudicating the corpus must not tailor cases to either policy's specific rule wording.

The Baseline C authoring team must not be given the DIO/ODI diagnostic mapping, its category labels, or the Challenger policy; otherwise, the generic condition may inherit the structure it is intended to test. Challenger authors may use the specified DIO/ODI mapping, but must not receive facts withheld from C. The same underlying domain facts must be available to both conditions. Authoring effort, available time, relevant ER/policy expertise, and non-policy execution resources should be matched as far as practicable; authors should not see the counterpart policy or its results before their versions are frozen, and adjudication rules should be prespecified. Prior DIO/ODI exposure must be recorded for all policy authors.

With one author or team per condition, observed differences are exploratory and cannot isolate diagnostic organization from team-specific expertise. For multi-team replication intended to support stronger comparative claims, independently constituted author teams should be allocated to conditions using a prespecified random or expertise-blocked procedure. The allocation procedure, expertise strata, prior DIO/ODI exposure, and any contamination or cross-condition exposure must be recorded. Repeated runs of one authored policy do not count as independent authoring replications. Stronger comparative claims require multiple independently constituted author teams per condition and an uncertainty analysis that reflects between-team variation. If such replication is infeasible, the result should be reported as a comparison of the particular authoring teams, not as an isolated effect of the DIO/ODI mapping.

The promotion process may itself govern whether recorded shared upstream origins count as independent corroboration. Strata whose principal challenge is that shared promotion rule should therefore be identified as common-control tests; they isolate an incremental mapping effect only if a distinct, prespecified diagnostic-to-constraint difference is also being tested. If Baseline C independently encodes the same constraints from the same facts, identical outcomes are expected and do not count as evidence of a DIO/ODI advantage. The intended difference is diagnostic organization and its explicit mapping to constraints, not additional information, validator capability, or execution resources.

The corpus should combine entity-resolution benchmarks with temporally evolving and ambiguous entities where licensing and annotation provenance permit reproducible use. Entity-overshadowing methods provide one source of inspiration [18], while TempEL provides a temporal-evolution setting [24]. Retrieval-augmented generation vulnerability research may inform future perturbation design [23]. The selected datasets, versions, licences, entity identifiers, ground-truth policies, case-construction procedure, and stratum weights must be documented in the experiment report.

The commit-boundary test set should contain: (i) cases where names, brands, acquisitions, or partnerships are insufficient to establish identity; (ii) positive identity-continuity cases supported by temporally applicable evidence; (iii) conflicting or stale evidence, including source-local identifier reassignment; and (iv) cases in which mention-to-identifier linkage is uncertain. A taxonomy-neutral stratum should additionally be defined from factual properties without using the DIO/ODI categories. Annotation must separately label (a) whether a mutation is permitted by the common predeclared evaluation reference and (b) whether it is referentially correct according to independent ground truth. Threshold selection and model configuration must be separated from the final test split. Entity-disjoint and perturbation-template-disjoint splits should be used where feasible.

Outcome adjudication for Safe Resolution Coverage (SRC), Unwarranted Abstention Rate (UAR), and False Positive Intervention Rate (FPIR) must use a common, independently specified evaluation reference, not either condition's authored operational policy. Correctness, safety, admissibility, resolvability, and benignness are adjudicated against that common reference by adjudicators independent of policy authors and masked to condition where practicable. PEVR is different: it measures compliance with each condition's applicable operational contract and is assessed against that contract. A policy-permitted mutation and a referentially correct mutation are separate labels; neither substitutes for the other.

Concurrency tests should use shared versioned evidence and graph state under controlled workloads. They should measure stale-authorization commits, version-conflict retries, uncertain-commit reconciliation, and whether any candidate is applied after its assumed graph, evidence content, authorization eligibility, or policy context has changed. The promotion process and its version must be held constant between Baseline C and the Challenger; outcome analysis should also distinguish policy specification, evidence promotion, conflict handling, and enforcement/version-control failures. Shared upstream-origin cases governed only by the common promotion contract should not be presented as evidence of a distinct DIO/ODI mapping effect.

\subsection{8.2 Evaluation Metrics}
Metrics should jointly report unsafe persistence, resolution coverage, abstention, and operational cost. A single safety metric is insufficient because a system that defers every case could appear safe while offering little useful resolution. All outcome labels below, except contract-compliance labels used by PEVR, are assigned against the common independent evaluation reference defined in Section 8.1.

\textbf{Counting codebook.} Before outcomes are examined, the protocol must define a case-level and operation-level counting codebook. A protected mutation attempt is identified by a unique \texttt{operation\_id}; retries, re-evaluations, and reconciliation attempts for the same intended operation do not create additional mutation attempts, while a genuinely new operation has a new identifier. The terminal outcome is assessed at a fixed observation/reconciliation cutoff. A case is resolved only when it reaches a correct terminal identity disposition within that window, including a correct non-match decision where non-identity is the ground-truth outcome. A temporary deferral later resolved within the window is not a final abstention; a case still unresolved at the cutoff remains unresolved. The codebook must define the ground-truth-positive, ground-truth-negative, benign, resolvable, admissible, and correctly resolved labels and their mapping to the case-level confusion matrix. Unknown transaction outcomes still unreconciled at the cutoff remain \texttt{unknown}; they must not be silently recoded as successful or failed commits.

\textbf{Policy-Enforcement Violation Rate (PEVR).} Report protected mutations committed despite a failed authorization condition, unresolved required conflict, invalid scope, or stale bound version, divided by all unique attempted protected mutations. Retries and re-evaluations under the same \texttt{operation\_id} count once in the denominator. This measures enforcement fidelity to the applicable operational contract; the raw numerator and denominator must also be reported. If there are no unique attempts, the rate is \texttt{undefined}.

\textbf{Referentially False Commit Rate (RFCR).} Report referentially false identity-changing commits, as determined by independent ground-truth annotation, divided by all committed identity-changing mutations. Also report the count of false commits per unique attempted protected mutation. This measures the combined effect of policy adequacy, promotion quality, and enforcement; it must not be interpreted as an enforcement-only metric. If no identity-changing mutation is committed, RFCR is \texttt{undefined}.

\textbf{Safe Resolution Coverage (SRC).}

\begin{center}
\begin{adjustbox}{max width=\linewidth}
$\displaystyle \operatorname{SRC}=\frac{|\text{Correctly and safely resolved cases under the common evaluation reference}|}{|\text{All cases in the prespecified evaluation scope}|}.$
\end{adjustbox}
\end{center}
The evaluation scope is fixed by the independent corpus and annotation criteria before system outputs are examined; it must not be defined by either supervisor's eligibility decisions. A resolved case must have a correct terminal match or non-match disposition and satisfy the common reference's safety criteria. Deferred and still-unknown cases remain in the denominator and do not count in the numerator. If the denominator is zero, SRC is \texttt{undefined}.

\textbf{Unwarranted Abstention Rate (UAR).}

\begin{center}
\begin{adjustbox}{max width=\linewidth}
$\displaystyle \operatorname{UAR}=\frac{|\text{Final unresolved outcomes on common-reference-resolvable cases}|}{|\text{All common-reference-resolvable cases}|}.$
\end{adjustbox}
\end{center}
A common-reference-resolvable case is one for which independent annotation confirms that sufficient admissible evidence exists within the specified observation window. Admissibility is judged against the common evaluation reference, not the condition's operational policy. An outcome is a final abstention only if it remains unresolved at the prespecified cutoff; temporary deferral followed by a correct resolution within the window is not counted as an abstention. If the denominator is zero, UAR is \texttt{undefined}.

\textbf{False Positive Intervention Rate (FPIR).}

\begin{center}
\begin{adjustbox}{max width=\linewidth}
$\displaystyle \operatorname{FPIR}=\frac{|\text{Unnecessary interventions on common-reference-benign and resolvable cases}|}{|\text{All common-reference-benign and resolvable cases}|}.$
\end{adjustbox}
\end{center}
A benign case is not automatically an intervention error merely because it contains uncertainty: the common reference must also establish that a safe terminal disposition was available within the observation window and that the halt/deferral was unnecessary under the prespecified criteria. Benignness, resolvability, and the intervention window are labeled independently before system comparison. If there are no cases meeting the denominator criteria, FPIR is \texttt{undefined}.

\textbf{Effectiveness.} Report Precision, Recall, and F1 under a prespecified case-level ground-truth policy and publish the confusion-matrix mapping. A predicted positive link between distinct referents is a false positive; a true positive link left unresolved at the cutoff despite being common-reference-resolvable counts as a false negative. Correct non-match decisions are negatives, not abstentions. Report unresolved and deferred outcomes separately. Precision, Recall, and F1 are each \texttt{undefined} when their respective denominator is zero; no zero value is substituted for an undefined metric.

\textbf{Commit integrity.} Report counts of stale-authorization commits, duplicate commits after retries, unreconciled \texttt{TIMEOUT\_UNKNOWN} outcomes, and commits that violate the bound target/effect scope. The expected value for protected stale-authorization commits is zero under the tested transaction contract; this is a testable requirement, not a claim already demonstrated. Report unreconciled unknown outcomes separately at the fixed cutoff, without classifying them as successes or failures.

\textbf{Efficiency.} The proposed Cost per Correct Resolution (CCR) is

\begin{center}
\begin{adjustbox}{max width=\linewidth}
$\displaystyle \operatorname{CCR}=\frac{\text{Total system cost}}{\text{Number of confirmed correctly resolved entities}}.$
\end{adjustbox}
\end{center}
Total cost should specify API/token pricing, human validation time and rates, and the compute-pricing basis over a common costing horizon. Policy-authoring and evaluation-corpus construction costs should be reported separately; the report must state whether either is included in the CCR numerator. The denominator must use a declared entity-deduplication rule and count only entities represented by correctly resolved cases. If no entities are correctly resolved, CCR is \texttt{undefined}; report it alongside SRC and abstention metrics.

Before outcomes are examined, the evaluation must prespecify task- and policy-family-specific decision margins: \(\delta_{\mathrm{SRC}}\) for the largest acceptable reduction in SRC when claiming non-inferior coverage, and \(\delta_{\mathrm{RFCR}}\) for the largest acceptable increase in RFCR when claiming non-inferior false-commit performance. Both margins must be nonnegative and justified from domain risk and coverage requirements rather than selected after observing results. The uncertainty procedure, interval level, analysis unit, case/stratum weights, and aggregation rule must also be prespecified. Because the success rule has two alternative routes, the inferential procedure must account for this two-route decision structure when calibrating uncertainty and the overall claim criterion (for example, through simultaneous bounds or another prespecified multiplicity-control approach). For multi-team inference, the primary comparison should reflect author-team variation; repeated runs of a single policy are nested observations, not independent authoring replications. Case, mutation, run, dataset split, and author team must not be treated as interchangeable units. No particular estimator is mandated in this conceptual paper, but the chosen procedure must respect the declared clustered/repeated structure.

Let \(L_S,U_S\) denote the prespecified uncertainty bounds for \(\Delta S=SRC_{Challenger}-SRC_C\), and \(L_R,U_R\) the corresponding bounds for \(\Delta R=RFCR_{Challenger}-RFCR_C\). The following are the \textbf{binding success criteria}, not examples. The incremental-value hypothesis is supported only if at least one of these two prespecified routes is established:

\begin{enumerate}
\item \textbf{Lower RFCR at non-inferior coverage:} \(L_S>-\delta_{\mathrm{SRC}}\) and \(U_R<0\).
\item \textbf{Improved coverage without a material RFCR increase:} \(L_S>0\) and \(U_R\leq\delta_{\mathrm{RFCR}}\).
\end{enumerate}
The second route supports improved coverage within the prespecified RFCR tolerance; it does not establish improved safety or strict Pareto improvement. If RFCR is undefined in either condition, neither RFCR-based comparative route is established; the affected route is non-evaluable, not evidence of safety superiority or inferiority. If neither route is established, the incremental-value hypothesis is not supported by the evaluation; this does not by itself establish equivalence, inferiority, or harm. A result in which Baseline C matches or Pareto-dominates the Challenger counts against incremental mapping value for the task and policy family tested, not against DIO/ODI as a broader framework. Component ablations should compare conventional rules, the generic authorization gate, and DIO/ODI-informed binding without bundling all controls into one intervention. Any evaluation of dependency-aware cascade mitigation belongs to future research and must be reported separately from this commit-boundary evaluation.

\section{9. Limitations and Future Research}
The single evidence-state version \(v_E\) is a deliberately conservative invalidation mechanism: a change to evidence unrelated to a particular \(q_m\) can invalidate a pending authorization and increase retries under contention. The design does not claim starvation-freedom or optimal concurrency. Because unresolved conflicts indicated by quarantined or \texttt{UNVERIFIED} items can defer an otherwise legitimate mutation, noisy or adversarial inputs may impede progress; abuse-resistant prioritization and adjudication are outside the present contract. Revoking evidence promotion invalidates pending authorizations, but does not retroactively undo completed commits; auditing dependent commits and deciding whether compensating mutations are warranted are outside this paper's scope. The proposed C-versus-Challenger comparison is also sensitive to policy-authoring skill, as discussed in Section 8.1; a single author or team per condition supports only an exploratory team-level comparison, while stronger claims about incremental mapping value require replication across independently constituted author teams.

\subsection{9.1 Dependency-Aware Evidence Flow}
The principal limitation is deliberate: the proposed boundary does not prevent an agent from retaining a mistaken hypothesis or from selecting later retrieval actions under its influence. A future dependency-aware architecture could track high-level derivation lineage among hypotheses, actions/queries, observations, and subsequent claims; distinguish multiple documents from genuinely independent corroboration; and expose material dependency information to later planning. Its key test would be whether it reduces unsafe mutation proposals or improves calibrated deferral under controlled hypothesis-conditioned perturbations, without causing disproportionate unwarranted abstention. The present paper does not specify an estimator or claim that such a layer is straightforward to implement.

\subsection{9.2 Domain-Policy Construction}
The proposal assumes that domain-configured identity-preserving constraints \(V_E\) can be supplied for an entity class and observation domain [6]. In closed enterprise environments, some constraints may be instantiated from schema or identity-management rules [12, 13]. In open-world settings, deciding which properties preserve identity across contexts and time remains difficult. Deriving defensible, domain-scoped constraints from unstructured and context-sensitive material remains an open problem; the present paper assumes that domain policy supplies these constraints and does not propose an automated extraction method.

\subsection{9.3 Multi-Agent and Contextual Negotiation}
Multi-agent environments may contain agents operating in different observation domains \(\Omega_A\neq\Omega_B\), producing legitimate context-specific reconstructions that differ without either necessarily being erroneous [6, 11, 17, 21]. A shared representation does not imply that every agent should produce an identical contextual reconstruction. Future work should develop protocols for sharing versioned evidence while preserving legitimate context-specific outputs and identifying genuine semantic fragmentation. The commit-boundary design addresses a class of persistence and write-contention risks; it does not resolve semantic disagreement among agents by itself.

\section{10. Conclusion}
Agentic ER enables iterative evidence acquisition and state updates, but an early identity hypothesis may influence later retrieval and create apparent confirmation. This paper deliberately does not claim to eliminate that feedback mechanism. It proposes a narrower architectural contribution: a DIO/ODI-informed boundary between evidence available to an agent for reasoning and evidence eligible to authorize a persistent identity-changing mutation.

The proposed binding function translates selected distinctions in referential, semantic, temporal, and contextual reconstruction failures into domain-configured constraints for particular mutation classes. Under the stated contract, a protected candidate may be committed only when the relevant candidate graph passes those constraints on claim-specific authorization-eligible evidence, the operation matches its authorized scope, and relevant state and policy versions remain valid. \texttt{UNVERIFIED} evidence is not positive authorization support unless separately promoted for the specific claim under the stated contract; insufficient support or unresolved material conflict leads to deferral rather than implicit permission.

This property is conditional on the adequacy of the domain policy, the claim-specific evidence-promotion process, and enforcement of the commit boundary. The invariant summarizes the specified obligations; it is not offered as a stand-alone theorem of semantic correctness. The contribution is the proposed DIO/ODI-informed organization of domain constraints for mutation-specific identity propositions, integrated with established requirements for claim-specific promotion, proposition coverage, conflict handling, authorized scope, and commit-time version integrity. The paper does not claim novelty for those contract mechanisms individually. It is an abstract design claim, not an empirical result or a guarantee that the underlying evidence is true. The evaluation protocol is proposed future work. A dependency-aware information-flow layer that addresses the upstream hypothesis-conditioned feedback loop remains an explicit research direction beyond the present commit-boundary contribution.

\section{References}
\begin{enumerate}
\item Altman, E. (1999). \emph{Constrained Markov Decision Processes}. Chapman \& Hall/CRC. \url{https://search.worldcat.org/title/41258998}
\item Ayoola, T., \& Fisher, J. (2022). Improving “entity linking” between texts and knowledge bases. \emph{Amazon Science}. \url{https://www.amazon.science/blog/improving-entity-linking-between-texts-and-knowledge-bases}
\item Barlaug, N. (2023). LEMON: Explainable Entity Matching. \emph{IEEE Transactions on Knowledge and Data Engineering}, 35(8), 8171–8184. \url{https://doi.org/10.1109/TKDE.2022.3200644}
\item Beránek, D. (2026a). \emph{The Crisis of Entity Resolution in Heterogeneous AI Environments: Data, Financial, and Semantic Dimensions} (Version 1.0.0). Zenodo. \url{https://doi.org/10.5281/zenodo.22257504}
\item Beránek, D. (2026b). \emph{Manifest of Digital Identity Optimization (DIO) and Ontology of Digital Identity (ODI): Canonical Multilingual Edition} (Version 1.0.0). Zenodo. \url{https://doi.org/10.5281/zenodo.21610934}
\item Beránek, D. (2026c). \emph{Minimum Viable Digital Identity Optimization (DIO): From Adversarial Reduction to the Semantic Field Dynamics of Digital Identity} (Version 1.0.0). Zenodo. \url{https://doi.org/10.5281/zenodo.23226205}
\item Beránek, D. (2026d). \emph{From Entity Resolution Anomalies to Reconstruction Failure Modes: A Taxonomic Note Following Minimum Viable Digital Identity Optimization (DIO)}. Zenodo. \url{https://doi.org/10.5281/zenodo.23269323}
\item Beránek, D. (2026). \emph{DIO/ODI Reference Vocabulary and Serialization} (Version 0.2.0). Zenodo. Concept DOI: \url{https://doi.org/10.5281/zenodo.23272897}
\item Beránek, D. (2026). \emph{DIO/ODI Reference Model} (Version 0.5.0) [Software release]. GitHub repository, release v0.5.0: \url{https://github.com/dan-beranek/dio-odi-reference-model/releases/tag/v0.5.0}. Archived software record: \url{https://doi.org/10.5281/zenodo.23643845}
\item Brunner, U., \& Stockinger, K. (2020). Entity Matching with Transformer Architectures—A Step Forward in Data Integration. In \emph{Proceedings of the 23rd International Conference on Extending Database Technology (EDBT)} (pp. 463–473). \url{https://doi.org/10.5441/002/edbt.2020.58}
\item Cemri, M., Pan, M. Z., Yang, S., Agrawal, L. A., Chopra, B., Tiwari, R., Keutzer, K., Parameswaran, A., Klein, D., Ramchandran, K., Zaharia, M., Gonzalez, J., \& Stoica, I. (2025). Why Do Multi-Agent LLM Systems Fail? \emph{Advances in Neural Information Processing Systems (NeurIPS 2025)}. \url{https://neurips.cc/virtual/2025/poster/121528}
\item Christophides, V., Efthymiou, V., Palpanas, T., Papadakis, G., \& Stefanidis, K. (2020). An Overview of End-to-End Entity Resolution for Big Data. \emph{ACM Computing Surveys}, 53(6), Article 127, 1–42. \url{https://doi.org/10.1145/3418896}
\item ISO/IEC. (2025). \emph{Information security, cybersecurity and privacy protection—A framework for identity management—Part 1: Core concepts and terminology} (ISO/IEC Standard No. 24760-1:2025). International Organization for Standardization. \url{https://www.iso.org/standard/24760-1}
\item Lee, J. D., \& See, K. A. (2004). Trust in Automation: Designing for Appropriate Reliance. \emph{Human Factors}, 46(1), 50–80. \url{https://doi.org/10.1518/hfes.46.1.50_30392}
\item Li, H., Li, S., Hao, F., Zhang, C. J., Song, Y., \& Chen, L. (2024). BoostER: Leveraging Large Language Models for Enhancing Entity Resolution. In \emph{Companion Proceedings of the ACM on Web Conference 2024 (WWW ’24)} (pp. 1043–1046). \url{https://doi.org/10.1145/3589335.3651245}
\item Li, Y., Li, J., Suhara, Y., Doan, A., \& Tan, W.-C. (2020). Deep Entity Matching with Pre-Trained Language Models. \emph{Proceedings of the VLDB Endowment}, 14(1), 50–60. \url{https://doi.org/10.14778/3421424.3421431}
\item Papadakis, G., Korovesis, P., Koubarakis, M., \& Palpanas, T. (2026). \emph{AgenticER: the next frontier in Entity Resolution}. arXiv:2607.27435. \url{https://doi.org/10.48550/arXiv.2607.27435}
\item Provatorova, V., Bhargav, S., Vakulenko, S., \& Kanoulas, E. (2021). Robustness Evaluation of Entity Disambiguation Using Prior Probes: the Case of Entity Overshadowing. In \emph{Proceedings of the 2021 Conference on Empirical Methods in Natural Language Processing (EMNLP 2021)} (pp. 10501–10510). \url{https://doi.org/10.18653/v1/2021.emnlp-main.820}
\item Talburt, J. R., \& Zhou, Y. (2015). \emph{Entity Information Life Cycle for Big Data: Master Data Management and Information Integration}. Morgan Kaufmann. \url{https://doi.org/10.1016/C2013-0-18748-X}
\item Wang, T., Chen, X., Lin, H., Chen, X., Han, X., Sun, L., Wang, H., \& Zeng, Z. (2025). Match, Compare, or Select? An Investigation of Large Language Models for Entity Matching. In \emph{Proceedings of the 31st International Conference on Computational Linguistics (COLING 2025)} (pp. 96–109). \url{https://aclanthology.org/2025.coling-main.8/}
\item Weick, K. E., Sutcliffe, K. M., \& Obstfeld, D. (2005). Organizing and the Process of Sensemaking. \emph{Organization Science}, 16(4), 409–421. \url{https://doi.org/10.1287/orsc.1050.0133}
\item Weng, S., Feng, Y., Zhang, J., Xie, X., Yu, J., \& Liu, J. (2026). \emph{ARGUS: Defending LLM Agents Against Context-Aware Prompt Injection}. arXiv preprint arXiv:2605.03378. \url{https://arxiv.org/abs/2605.03378}
\item Xian, X., Wang, G., Bi, X., Zhang, R., Srinivasa, J., Kundu, A., Fleming, C., Hong, M., \& Ding, J. (2025). On the Vulnerability of Applying Retrieval-Augmented Generation within Knowledge-Intensive Application Domains. In \emph{Proceedings of the 42nd International Conference on Machine Learning} (Proceedings of Machine Learning Research, 267, pp. 68292–68315). \url{https://proceedings.mlr.press/v267/xian25a.html}
\item Zaporojets, K., Kaffee, L.-A., Deleu, J., Demeester, T., Develder, C., \& Augenstein, I. (2022). TempEL: Linking Dynamically Evolving and Newly Emerging Entities. \emph{Advances in Neural Information Processing Systems (NeurIPS 2022)}, 35. \url{https://doi.org/10.52202/068431-0135}
\item Zeakis, A., Papadakis, G., Skoutas, D., \& Koubarakis, M. (2023). Pre-trained Embeddings for Entity Resolution: An Experimental Analysis. \emph{Proceedings of the VLDB Endowment}, 16(9), 2225–2238. \url{https://doi.org/10.14778/3598581.3598594}
\item ISO. (2020). \emph{Financial services—Legal entity identifier (LEI)—Part 1: Assignment} (ISO 17442-1:2020). International Organization for Standardization. \url{https://www.iso.org/standard/78829.html}
\item Global Legal Entity Identifier Foundation (GLEIF). (n.d.). \emph{The Legal Entity Identifier (LEI)}. \url{https://www.gleif.org/en/organizational-identity/lei-vlei/the-legal-entity-identifier-lei}
\item Zhu, G., \& Wang, C. (2026). \emph{Runtime Authorization of Self-Generated Subgoals in Long-Horizon Tool-Using AI Agents}. arXiv preprint arXiv:2610.04975. \url{https://arxiv.org/abs/2610.04975}
\item Zhu, H., Liang, J., Hou, M., Tang, R., Zhu, X., Yang, J., Mao, Y., \& Wu, F. (2026). \emph{From Business Events to Auditable Decisions: Ontology-Governed Graph Simulation for Enterprise AI}. arXiv preprint arXiv:2604.08603. \url{https://arxiv.org/abs/2604.08603}
\item World Wide Web Consortium (W3C). (2017). \emph{Shapes Constraint Language (SHACL)}. W3C Recommendation. \url{https://www.w3.org/TR/shacl/}
\item World Wide Web Consortium (W3C). (2013). \emph{PROV-O: The PROV Ontology}. W3C Recommendation. \url{https://www.w3.org/TR/prov-o/}
\item Santos-Grueiro, I. (2026). \emph{Temporary Authority, Permanent Effects: Commit-Time Authorization for LLM Agents}. arXiv preprint arXiv:2607.10487. \url{https://arxiv.org/abs/2607.10487}
\item Aldinucci, M., Bracciali, A., Colonnelli, I., Medić, D., Mulone, A., Roversi, L., \& Santimaria, M. E. (2026). Hard-KLAIM: Hardening MCP-Enabled Agentic Workflows with a Locality-Aware Type-and-Effect System for Trace Safety. \emph{Journal of Logical and Algebraic Methods in Programming}, Article 101191. \url{https://doi.org/10.1016/j.jlamp.2026.101191}
\item Zhang, X., Cheng, Z., Liu, K., Sun, Z., Fan, S., Yuan, T., Yuan, B., Qiang, W., \& Zou, D. (2026). \emph{AGATE: Provenance-Based Runtime Defense Against Compositional Attacks on LLM Agents}. arXiv preprint arXiv:2609.30830. \url{https://arxiv.org/abs/2609.30830}
\item Wu, M., Wang, L., Zhang, Y., \& Deng, J. (2026). \emph{From Intent to Execution Grant: An Execution-Boundary Conformance Profile for High-Risk AI Actions} (EBL-Core). arXiv preprint arXiv:2609.11596. \url{https://arxiv.org/abs/2609.11596}
\item Zheng, C., \& Yang, G. (2026). \emph{CONTINUITY: Security-Context Contracts for Composable LLM Agent Controls}. arXiv preprint arXiv:2609.05269. \url{https://arxiv.org/abs/2609.05269}
\item Arasu, A., Ré, C., \& Suciu, D. (2009). Large-Scale Deduplication with Constraints Using Dedupalog. In \emph{2009 IEEE 25th International Conference on Data Engineering} (pp. 952–963). \url{https://doi.org/10.1109/ICDE.2009.43}.
\end{enumerate}
\newpage

\section{Methodological Note on Model-Assisted Development of This Paper}
This paper was developed through a human-led, model-assisted process of conceptual exploration, architectural formulation, source-oriented research, manuscript drafting, and adversarial review. The author supplied the DIO/ODI research context and directed the problem framing, made the consequential decision to focus the present contribution on a commit-boundary authorization guardrail, and kept broader dependency-aware control of the hypothesis-conditioned cascade as a direction for future research. The architecture and manuscript emerged through iterative interaction rather than from a single model-generated output.

Generative AI systems contributed substantively, not only as language-editing tools. GPT-based systems supported later-stage synthesis, drafting, formalization, and revision of the architecture, interface contracts, algorithms, and prose. Claude contributed design criticism and adversarial review. Gemini was used during an earlier exploratory phase; material from that phase was subsequently reassessed rather than treated as authoritative. Perplexity supported source-oriented research, bibliographic checks, and targeted regression review, while Qwen and Kimi were used for adversarial critique of selected drafts. The roles and versions of the systems changed during development, and the work was not a controlled comparison of model capabilities. The v48 revision received separate, task-specific model-assisted adversarial reviews from Claude, Gemini, Perplexity, Lumo, and Mistral; where a service interface did not expose an exact model release, no specific version is asserted. The manuscript underwent iterative, versioned revision, including reviews aimed at exposing counterexamples, unsupported novelty claims, citation problems, and inconsistencies between the stated invariant and its algorithmic realization.

Model outputs were treated as fallible proposals and critiques, not as evidence of system performance or as independent peer review. The author selected, rejected, and revised suggestions and remains responsible for the paper's framing, design decisions, interpretations, claims, references, and final text. The author also retains responsibility for verifying retained factual claims and bibliographic details against their underlying sources. No empirical results are reported on the basis of model outputs; the evaluation protocol remains proposed future work. The paper is a model-assisted conceptual proposal developed under the author’s direction. Its stated properties remain conditional on the assumptions and limitations described above; it is neither implemented nor empirically validated.

\end{document}